\documentclass{article}

\usepackage[english]{babel}

\usepackage[letterpaper,top=2cm,bottom=2cm,left=2.25cm,right=2.25cm,marginparwidth=1.75cm]{geometry}

\usepackage{xcolor} 
\usepackage{mdframed} 
\usepackage{listings} 

\usepackage{amsmath} 
\usepackage{amsfonts}
\usepackage{graphicx}
\usepackage{listings}
\usepackage{booktabs}
\usepackage{parskip}
\usepackage{comment}
\usepackage{mdframed}
\usepackage{wrapfig}
\usepackage{threeparttable}
\usepackage{float}
\usepackage{csquotes}
\usepackage{multirow}
\usepackage{geometry}
\usepackage{array}
\usepackage{subcaption}
\usepackage{booktabs}
\usepackage[most]{tcolorbox}
\usepackage{wrapfig}
\usepackage{adjustbox}
\usepackage{graphicx} 
\usepackage{caption} 
\usepackage{subcaption} 

\usepackage[colorlinks=true, allcolors=blue]{hyperref}
\usepackage[
backend=biber,
style=authoryear,
]{biblatex}

\usepackage[colorlinks=true, allcolors=blue]{hyperref}

\def\sym#1{\ifmmode^{#1}\else\(^{#1}\)\fi}

\title{An Empirical Study of Scaling Laws for Transfer}
\author{Matthew Barnett\\
    Epoch AI\\
    \texttt{matthew@epochai.org}
}
\date{August 2024}

\begin{document}

\maketitle

\begin{abstract}
We present a limited empirical study of scaling laws for transfer learning in transformer models. More specifically, we examine a scaling law that incorporates a ``transfer gap" term, indicating the effectiveness of pre-training on one distribution when optimizing for downstream performance on another distribution. When the transfer gap is low, pre-training is a cost-effective strategy for improving downstream performance. Conversely, when the gap is high, collecting high-quality fine-tuning data becomes relatively more cost-effective. Fitting the scaling law to experiments from diverse datasets reveals significant variations in the transfer gap across distributions. In theory, the scaling law can inform optimal data allocation strategies and highlights how the scarcity of downstream data can bottleneck performance. Our findings contribute to a principled way to measure transfer learning efficiency and understand how data availability affects capabilities.
\end{abstract}

\section{Introduction}
In recent years, a number of papers have uncovered machine learning scaling laws---defined as empirical regularities that describe how the performance of a model increases as a function of scale, usually in parameter count and data (\cite{hestness2017deep}, \cite{kaplan2020scaling}, \cite{hoffmann2022training}). \cite{hernandez2021scaling} described scaling laws for transfer learning, showing how the transfer learning properties of models change as a function of model size. The primary result was that the degree of transfer---as measured by the amount of effective data transferred from one distribution to another---follows a simple power law in parameter count and fine-tuning data size. However, their analysis left much room for further exploration, as it only considered transfer learning from English to Python, and did not explore the relationship between the pre-training data size and the degree of downstream transfer learning.

Scaling laws for transfer are important to study because they inform the degree to which progress in machine learning is bottlenecked by data for specific tasks. Consider that to achieve high performance on some tasks, one standard approach in the foundation model paradigm is to pre-train a model on a large, diverse distribution and then fine-tune it on a particular downstream task (\cite{bommasani2022opportunities}). An alternative approach to fine-tuning for adapting pre-trained language models to downstream tasks is in-context learning, where a model is given a sequence of demonstrations within a prompt (\cite{brown2020language}, \cite{mosbach2023fewshot}). The effectiveness of these approaches depends on the degree of knowledge transfer from the pre-trained model to the downstream task or distribution. If the degree of transfer is high, then the cost of fine-tuning is typically low, allowing foundation models to efficiently learn to perform well on downstream tasks with minimal fine-tuning data. Conversely, if the degree of transfer is low, then performance is more heavily dependant on the amount of high-quality fine-tuning data available, motivating the curation of large fine-tuning datasets.

To help study whether high-quality fine-tuning data might bottleneck performance on important tasks, we devised a simple framework for measuring the transfer gap between two distributions using scaling laws for transfer. In this article, we loosely define the transfer gap as the maximum theoretical benefit of pre-training, achievable in the limit of infinite pre-training data. More specifically, the transfer gap is a quantity that emerges from an empirical scaling law for transfer when evaluating the limit of the scaling law as the pre-training variable tends towards infinity. By empirically comparing the predictive validity of various functional forms, we have settled on the following scaling law for transfer, where $p$ represents the number of pre-training steps and $f$ represents the number of fine-tuning data points.

\begin{equation}
L(p, f) = \left(\underbrace{A \cdot p^{-\alpha}}_{\text{Pre-training term}} + \underbrace{G}_{\text{Transfer gap}}\right) \cdot \underbrace{f^{-\beta}}_{\text{Fine-tuning term}} + \underbrace{E}_{\text{Irreducible loss}}
\end{equation}

In this scaling law, taken from \cite{mikami2021scaling}, the transfer gap determines the efficiency of fine-tuning in the limit of infinite pre-training data. By measuring the transfer gap from inexpensive and abundant pre-training distributions to costly downstream distributions, we can predict how much pre-training might improve performance on a given downstream distribution. Thus, measuring the transfer gap across various tasks can meaningfully inform the relative difficulty of automating downstream tasks, although drawing firm conclusions about the difficulty of automation is premature given the limited data in this study. 

Our study relies heavily on the Pythia model suite (\cite{biderman2023pythia}), providing a large set of transformer models trained on the Pile at various model sizes and checkpoints during training. We fine-tune a 2.8 billion parameter model on various downstream language datasets, including a mathematics dataset, a genetic sequence dataset, a statistics textbook, and a synthetic dataset of fictional biographies within a procedurally generated universe. After fitting the scaling law to empirical data, we demonstrate how this scaling law can be used to answer concrete questions, such as ``How valuable is it to collect more fine-tuning data for a given task, compared to scaling up pre-training data?" and ``What is the transfer gap from one distribution to another?".

Our analysis is incomplete for several reasons. Firstly, we do not derive a scaling law in model size.\footnote{In an initial version of this study, we attempted to include model size as an independent variable in our experiments. Nonetheless, we concluded that it was too costly to generate a sufficient number of data points to reliably fit a scaling law in three independent variables, given our limited compute budget. As a result, in the present study, we focus instead on a scaling law in pre-training steps and fine-tuning data points only.} Secondly, our results are largely limited to language datasets and were obtained with a relatively modest compute budget. For these reasons, we cannot necessarily extrapolate these results to much larger model sizes, of the type that exist at the current commercial frontier (e.g. \cite{openai2024gpt4}, \cite{geminiteam2024gemini}). Nonetheless, we hope our study can serve as a limited foundation for further research, stimulating more investigations into empirical scaling laws for transfer learning, enhancing our understanding of the role of data in advancing AI progress, and providing deeper insights into the difficulty of automating various economically important tasks.

\section{Related work}
\textbf{Neural scaling laws} The modern study of neural scaling laws began with \cite{hestness2017deep}, who observed a power law relationship between data size and error across image, language, and speech domains. Similarly, \cite{rosenfeld2019constructive} identified a power law relationship involving both model size and data. The work of \cite{kaplan2020scaling} and \cite{hoffmann2022training} furthered this research in language modeling, emphasizing the implications for allocating compute budgets—a tradition we continue in this study.

\cite{hernandez2021scaling} studied scaling laws for transfer in model size and fine-tuning data. Unlike their study, ours differs in three key ways. First, our study examines pre-training steps and fine-tuning data. Second, rather than studying the effective data transferred, our study directly investigates how test loss varies as a function of training inputs. These distinctions are crucial because they allow us to directly estimate the ``transfer gap" between two distributions, enabling a more fine-grained analysis of the limits to pre-training. Finally, we examine transfer across multiple diverse datasets, not just from English to Python.

Another similar study is \cite{mikami2021scaling}, which investigated how pre-training data affects transfer properties, similar to our approach. Unlike their study, ours involved training models to convergence on the fine-tuning data but not on the pre-training data, which we believe is a more realistic assumption in the large-data regime. Their research also focused on image classification, whereas ours is focused on natural language processing. Moreover, we aim to measure the ``transfer gap" between various distributions more directly and show how these gaps inform economic trade-offs when training large machine learning models, an aspect not addressed by \cite{mikami2021scaling}. Nonetheless, we borrow the scaling law form from \cite{mikami2021scaling}, which they postulated through theoretical analysis. For convenience and transparency, we detail how this scaling law can be derived in \autoref{app:derive_scaling_law}.

While our study focuses on transfer learning scaling laws between language datasets, \cite{aghajanyan2023scaling} investigated scaling laws for generative language models trained on multiple modalities. They modeled the individual contributions and interactions between modalities, identifying regimes of competition and synergy, where training on multiple modalities either negatively or positively affected performance, respectively. Although their study analyzed a distinct scenario—training on multiple modalities during pre-training, rather than pre-training on one dataset and fine-tuning to convergence on another—their study complements ours through the use of scaling laws across different datasets. The interaction terms in \cite{aghajanyan2023scaling}'s scaling law suggest transfer between datasets, which we attempt to quantify precisely using our framework.

In contrast to this study, \cite{zhang2024scaling} compared full-model tuning (FMT) with parameter-efficient tuning (PET). Their findings from experiments on bilingual LLMs suggested a multiplicative scaling law between fine-tuning data and other factors, indicating that model scaling offers more benefits than scaling pre-training data, with PET showing limited effectiveness. By comparison, our study exclusively employs full-parameter tuning for all experiments.

A review of the neural scaling law literature can be found in \cite{epoch2023scalinglawsliteraturereview}. The scarcity of previous work on scaling laws for transfer learning highlighted in their review further accentuates the need for research in this area.

\textbf{Transfer learning} Broadly speaking, the concept of transfer learning has been ubiquitous in machine learning. Most relevant to this study is its function as the key concept underpinning foundation models (\cite{bommasani2022opportunities}). Transfer learning involves taking knowledge learned on one task and applying it to another. This approach has been successfully used to build highly versatile language models that are pre-trained on large corpora and fine-tuned to perform well on various downstream tasks, with examples including \cite{openai2024gpt4}, \cite{geminiteam2023gemini}, and \cite{jiang2024mixtral}. An early exploration of this concept was provided by \cite{raffel2023exploring}, which informed subsequent work.

Theoretical investigations of transfer learning have been explored by \cite{Baxter_2000}, \cite{maurer2016benefit}, and \cite{tripuraneni2020theory}. These papers generally attempt to explain transfer learning by appealing to the concept of representation learning. In effect, when a model learns a representation from a pre-training distribution shared by downstream tasks, it enables the model to learn these tasks with fewer data points. While these studies provide valuable insights into the mechanisms behind transfer learning, it is challenging to directly apply these insights to derive a scaling law for transfer learning in a manner that would be most directly applicable to the present study. Nevertheless, these studies are useful for analyzing the convergence behavior of models in a transfer learning setting. For example, \cite{maurer2016benefit} proposes a framework for bounding the convergence rate of a learning algorithm depending on the number of samples from the pre-training task ($p$) and the downstream task ($f$), which follows a power law of the form $O(p^{-1/2}) + O(f^{-1/2})$.

\section{The scaling law for transfer}
\label{app:scaling_laws}
We investigate scaling laws for transfer learning using a 2.8 billion parameter transformer model from the Pythia suite of models (\cite{biderman2023pythia}). Specifically, we attempt to fit a scaling law to empirical data that relates cross-entropy loss on the fine-tuning distribution to the number of pre-training tokens seen, denoted \(p\), and the size of fine-tuning data, denoted \(f\). In this article, we refer to this mathematical relationship as a \textbf{scaling law for transfer}.

Considering a number of potential scaling law forms, we identified one that was both simple and appeared to perform well according to empirical tests. This form was taken directly from \cite{mikami2021scaling}, and serves as the basis of the analysis presented here:
\begin{equation}
L(p, f) = (A \cdot p^{-\alpha} + G) \cdot f^{-\beta} + E
\end{equation}
This scaling law is composed of four distinct terms. The first term, \(A \cdot p^{-\alpha}\), is the pre-training term: a power law in pre-training data steps. The second term is the transfer gap, \(G\). The utility of the transfer gap can be seen by considering \(\lim_{p\to\infty} L(p, f)\). In the infinite pre-training regime, the only value remaining inside the parentheses is the transfer gap, implicitly setting the ultimate efficiency of transfer learning from the pre-training distribution to the fine-tuning distribution. The third term, \(f^{-\beta}\), is the fine-tuning term, which is a power law. Finally, the fourth term, \(E\), represents the intrinsic entropy of the fine-tuning distribution, setting the maximum theoretical limit of training performance.

We compare this scaling law to the scaling law proposed by \cite{hernandez2021scaling} and discuss how it can be derived in \autoref{app:derive_scaling_law}. It is important to note that there are a large number of potential forms that we did not consider. Consequently, it is highly plausible that different functional forms for the scaling law might better predict the data. Several lines of research have aimed at uncovering the theoretical underpinnings of scaling laws in machine learning (see, for example, \cite{sharma2020neural} and \cite{bahri2021explaining}), but as of this writing, we do not have sufficient confidence in any theoretical framework to definitively derive the ``correct" form of the scaling law for transfer in our analysis. For these reasons, we rely more on empirical tests to demonstrate the robustness of fit.

To compare various potential scaling law forms against each other, we employ a form of cross-validation as described in \autoref{model_selection} that enables us to test how well the model predicts data out-of-distribution. We also compute goodness-of-fit statistics using bootstrapping.

\section{Methods}
\subsection{Datasets}
To empirically measure the transfer learning properties of transformer models, we fine-tuned them on five datasets: four diverse language datasets and one biological sequence dataset. One of the language datasets was a synthetic dataset designed to be subtly different from any natural dataset, to test whether the resulting scaling law for transfer uncovered genuine transfer to novel distributions. The other three language datasets were naturally collected. Two of the language datasets (the math arXiv and statistics textbook datasets) contain data that was published after the creation of the Pile, and thus are highly unlikely to be contained in the pre-training dataset. By contrast, the Enron emails dataset (\cite{enron_dataset}) is documented to have been included in the Pile (\cite{biderman2023pythia}). Finally, the genomic dataset (\cite{felis_catus_9_ensemble}) was selected to determine whether we could uncover the transfer learning gap to very distant distributions. More details for all of these datasets are outlined in Table~\ref{tab:datasets}. A longer description and sample from each dataset can be found in \autoref{app:dataset_details}.

\begin{table}[h]
\centering
\begin{tabular}{lllp{6cm}}
\toprule
\textbf{Dataset Name} & \textbf{Source} & \textbf{Type} & \textbf{Description} \\
\midrule
Fictional encyclopedia & Synthetic & Natural language & Generated by GPT-3.5 to simulate biographies in a fictional universe, specifically designed to test scaling laws for transformer models. \\
Math arXiv & Collected & Natural language & Consists of recent mathematics papers from the arXiv database, parsed in LaTeX format (uploaded after knowledge cutoff). \\
Statistics textbook & Collected & Natural language & Derived from ``Statistics for Ecologists" focusing on modern regression methods, includes both frequentist and Bayesian approaches, open-source. \\
Enron emails & Collected & Natural language & Comprises a subset of the well-known Enron email dataset, providing a rich source of real-world corporate email communication data. \\
House cat genome & Collected & Genomic data & Consists of the sequenced genome of a domestic cat, used for biological and genetic research. \\
\bottomrule
\end{tabular}
\caption{Datasets used for fine-tuning the large language models}
\label{tab:datasets}
\end{table}

\subsection{Training}
All models were part of the Pythia suite of transformer models, pre-trained on the Pile (\cite{gao2020pile}, \cite{biderman2023pythia}). Specifically, we selected 15 checkpoints during the training of a 2.8 billion parameter model from the Pythia suite. These checkpoints were saved at various stages during training, with the final checkpoint occurring at 143,000 steps and a batch size of 2,097,152 tokens. For each of these 15 checkpoints, and across each dataset, we conducted 15 experiments by fine-tuning the model to convergence on subsets of the datasets, with sizes ranging from 10 to 1,100 tokens and a context length of 256, employing full-parameter fine-tuning (\cite{raffel2023exploring}). 

We trained to convergence on the fine-tuning data because we believe this procedure captures the intuitive idea of learning as much as possible from limited fine-tuning data, which we believe is most relevant for studying how to optimize performance when downstream data is scarce. The number of pre-training steps and number of fine-tuning data points for all of the experiments are summarized in \autoref{tab:experiments}. During fine-tuning, we update all model parameters and did not explore parameter-efficient fine-tuning methods such as LoRA (\cite{hu2021lora}), which we suggest as a direction for future research.

The models were trained on an H100 GPU, utilizing a batch size of 10, and employing gradient accumulation steps of 25. The optimizer of choice was AdamW, configured with a learning rate of $1 \times 10^{-5}$, betas of 0.9 and 0.999, an epsilon value of $1 \times 10^{-8}$, and no weight decay. These models were trained to convergence on each subset of every fine-tuning dataset with an early stopping patience of 3, resulting in a total of 750 training runs across all of our experiments.

\subsection{Scaling law evaluation}
\label{model_selection}
To better understand the robustness of the scaling law, we conducted a form of step-wise cross-validation on the data to test its out-of-distribution predictive abilities. The step-wise cross-validation process involves selecting a set of thresholds for each input variable: pre-training tokens seen, and fine-tuning data size. All possible combinations of these thresholds are then generated. For each combination, the data is split into a training set containing data points below the threshold values and a testing set containing data points above the threshold values. The scaling law is then fitted to the training set and evaluated on the testing set, assessing how well it predicts data points that were not included in the training phase.

By iterating over all combinations of thresholds and assessing the scaling law's performance on data points outside the range used for fitting, this process helps to assess the model's ability to generalize and make accurate predictions about language models trained with more pre-training steps or more fine-tuning data points. This step-wise cross-validation process is outlined in Appendix~\ref{model_selection_details} and is visually summarized in \autoref{fig:Plots/updated_cross_validation}. Additionally, for each general scaling law, we employ L2 regularization, with a set of different hyperparameter choices for each scaling law variant. More details about the choice of regularization hyperparameters are provided in Appendix~\ref{model_selection_details}.

\section{Results}
Here, we provide an overview of some key general results from the empirical study. However, an in-depth presentation of the experimental data can be found in \autoref{data_presentation_results}.

\subsection{Pre-training acts to reduce loss relatively independently from the transfer gap}

\autoref{tab:fitted_parameters} presents the empirically fitted parameter values of the scaling law for transfer on each dataset. Recall that the key equation we fit is:
\begin{equation*}
L(p, f) = \left(\underbrace{A \cdot p^{-\alpha}}_{\text{Pre-training term}} + \underbrace{G}_{\text{Transfer gap}}\right) \cdot \underbrace{f^{-\beta}}_{\text{Fine-tuning term}} + \underbrace{E}_{\text{Irreducible loss}} \tag{1}
\end{equation*}

Comparing these values across datasets reveals some interesting patterns. The exponent in the pre-training term, $\alpha$, exhibited relatively low variation compared to the other terms, across the different fine-tuning datasets. More specifically, the coefficient of variation for the pre-training exponent, $\alpha$, was found to be less than 25\% of that for the fine-tuning exponent, $\beta$, across the fine-tuning datasets. The coefficients of variation across datasets for all parameters are shown in \autoref{tab:fitted_parameters}.

These results suggest a consistent contribution to the reduction of loss from pre-training that occurs independently of the transfer gap $G$. Since the degree of downstream transfer is mediated by the pre-training and transfer gap terms, this result strengthens the interpretation of the transfer gap as representing the directed proximity between two different distributions. 

\begin{table}[htbp]
    \centering
    \caption{Fitted parameter values for the fine-tuning datasets, with standard errors shown in parentheses}
    \label{tab:fitted_parameters}
    \begin{adjustbox}{max width=\textwidth}
    \begin{tabular}{@{}lccccc@{}}
        \toprule
        Dataset & \(A\) & \(G\) & \(\alpha\) & \(\beta\) & \(E\) \\
        \midrule
        Fictional encyclopedia & 284.766 (38.55) & 2.570 (0.18) & 0.730 (0.02) & 0.123 (0.01) & 0.538 (0.19) \\
        Math arXiv & 317.966 (29.31) & 0.166 (0.04) & 0.756 (0.02) & 0.059 (0.01) & 1.758 (0.04) \\
        Statistics textbook & 177.321 (20.93) & 1.305 (0.26) & 0.627 (0.02) & 0.126 (0.02) & 1.367 (0.19) \\
        Enron emails & 181.482 (19.32) & 0.595 (0.11) & 0.611 (0.02) & 0.159 (0.01) & 1.373 (0.07) \\
        House cat genome & 43.556 (7.85) & 0.548 (0.02) & 0.718 (0.05) & 0.228 (0.03) & 2.677 (0.04) \\
        \midrule
        Coefficient of variation & 0.528 & 0.851 & 0.094 & 0.432 & 0.490 \\
        \bottomrule
    \end{tabular}
    \end{adjustbox}
\end{table}

Contrary to our initial expectations, the transfer gap for the house cat genome dataset was relatively small, estimated to be only $0.548$, despite the fact that it was not a natural language dataset, and thus should intuitively be expected to have a low degree of transfer (and thus a large transfer gap). However, we believe this result can be reconciled with our intuitions by considering the relatively high estimated irreducible loss for the house cat genome dataset, at $2.677$, indicating that the dataset has high intrinsic entropy, and relatively little learnable structure.

Given the interpretation of the transfer gap, $G$, as indicating the maximum theoretical benefit of pre-training, it is unsurprising that this quantity would be low when fine-tuning on a dataset with little learnable structure. In a trivial case, fine-tuning on a dataset of random digits would likely exhibit a high degree of transfer learning, since the model only needs to learn the general pattern of random digit generation in order to achieve optimal downstream performance. It seems likely that such a simple underlying structure in the fine-tuning data could be easily picked up in the limit of maximum pre-training.

\subsection{The scaling law for transfer can be efficiently estimated using relatively little compute}

Despite only utilizing 150 data points per dataset, we have obtained a fairly robust fit for the scaling law on each dataset, as indicated by the relatively narrow confidence intervals for each parameter. Specifically, the standard error for the exponent parameters generally ranged between $0.01$ and $0.03$. By comparison, the standard errors for the exponent parameters in the scaling law in \cite{hoffmann2022training} was around $0.02$, indicating a similar degree of uncertainty (\cite{besiroglu2024chinchilla}). These intervals were derived from bootstrapping with 4,000 samples per dataset, and the results are shown in \autoref{tab:bootstrap_confidence_intervals}. The standard errors for all of the parameters are detailed in \autoref{tab:fitted_parameters}.

To estimate the necessary compute to run our full set of experiments, we can employ the heuristic that \(\text{Compute}={6\cdot N \cdot D}\) where \(N\) is the number of parameters in the model, and \(D\) is the number of tokens seen during a single epoch (\cite{kaplan2020scaling}, \cite{hoffmann2022training}). Since there were 750 training runs in total, we can use the following formula:

\begin{equation}
    \text{Estimated compute}=6 \cdot \sum_{t \in T} (\text{Epochs until convergence for $t$}) \cdot N \cdot D_t
\end{equation}
where $T$ represents the set of all training runs. Using this formula, we get $\text{total estimated compute }\approx 4.77\cdot 10^{16} \text{ FLOP}$. To put this number in perspective, it is less than one millionth the estimated compute used to train Meta's Llama 3 70B (\cite{llama3modelcard}, \cite{epoch2023pcdtrends}). This estimate underscores the potential to obtain better estimates of the parameters in scaling laws for transfer using very little compute. These laws can, in turn, can inform data allocation strategies for larger training runs.
\begin{table}[htbp]
    \centering
    \caption{95\% Confidence intervals for bootstrapped parameters}
    \label{tab:bootstrap_confidence_intervals}
    \begin{adjustbox}{max width=\textwidth}
    \begin{tabular}{@{}lccccc@{}}
        \toprule
        Dataset & \(A\) & \(G\) & \(\alpha\) & \(\beta\) & \(E\) \\
        \midrule
        Fictional encyclopedia & [227.69, 378.80] & [2.34, 3.05] & [0.684, 0.781] & [0.096, 0.151] & [0.05, 0.81] \\
        Math arXiv & [258.27, 373.16] & [0.00, 0.17] & [0.719, 0.788] & [0.038, 0.070] & [1.72, 1.89] \\
        Statistics textbook & [132.82, 214.86] & [0.95, 1.98] & [0.594, 0.658] & [0.086, 0.148] & [0.83, 1.59] \\
        Enron emails & [129.67, 205.41] & [0.44, 0.86] & [0.576, 0.641] & [0.118, 0.176] & [1.18, 1.46] \\
        House cat genome & [13.41, 44.17] & [0.52, 0.60] & [0.567, 0.780] & [0.148, 0.274] & [2.58, 2.72] \\
        \bottomrule
    \end{tabular}
    \end{adjustbox}
\end{table}

\section{Discussion}
\subsection{Trading off pre-training and fine-tuning data collection}
\label{app:discussion_tradeoffs}

\begin{figure}[!htbp]
    \centering \includegraphics[width=0.8\textwidth]{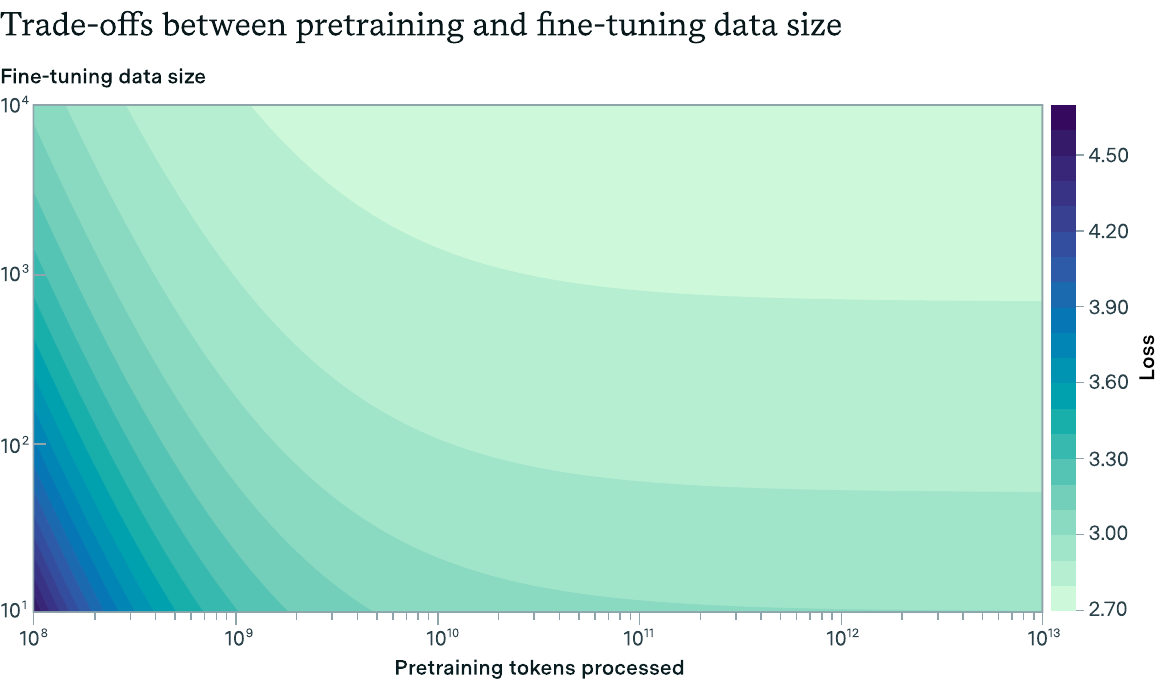}
    \caption{This plot presents the trade-offs between expanding pre-training and collecting more fine-tuning data to achieve low loss on the synthetic fictional encyclopedia dataset. The isolines, or lines of equivalent loss values, delineate the points at which equal loss is achievable given different combinations of pre-training steps and fine-tuning data points. The plot demonstrates that at low pre-training values, significant benefits can be gained from both increasing fine-tuning data and expanding pre-training. Conversely, at high pre-training values, the marginal benefit of additional pre-training diminishes, making the collection of more fine-tuning data points increasingly valuable for reducing loss.}
    \label{fig:Plots/iso_loss_lines}
\end{figure}

Consider a situation where there are costs associated with fine-tuning data collection, and we have a fixed budget, which can be allocated to either scaling pre-training or collecting more fine-tuning data. By utilizing the scaling law for transfer, we can determine under which circumstances it is valuable to collect more data or scale pre-training.

Consider the simplest case where one is optimizing for performance on only a single downstream distribution. Let \(B\) represent the budget, \(C_p\) the cost of a single pre-training step, and \(C_f\) the cost of collecting a fine-tuning datapoint and training on it (all in dollars). Given this setup, the optimal allocation of spending on pre-training and fine-tuning for a model of fixed size is given by the solution to the following optimization problem:

\begin{equation}
    \begin{aligned}
        & \underset{p, f}{\text{minimize}}
        & & L(p, f) \\
        & \text{subject to}
        & & C_p \cdot p + C_f \cdot f \leq B,
    \end{aligned}
\end{equation}

\begin{figure}[ht]
    \centering
    \begin{subfigure}[b]{0.49\textwidth}
        \centering
        \includegraphics[width=\textwidth]{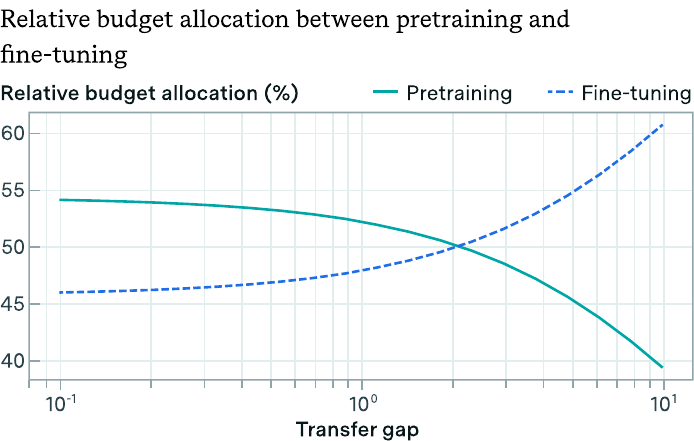}
        \caption{Budget allocation vs. transfer gap \( G \).}
        \label{fig:evolution_of_fine_tuning}
    \end{subfigure}%
    \hfill
    \begin{subfigure}[b]{0.49\textwidth}
        \centering
        \includegraphics[width=\textwidth]{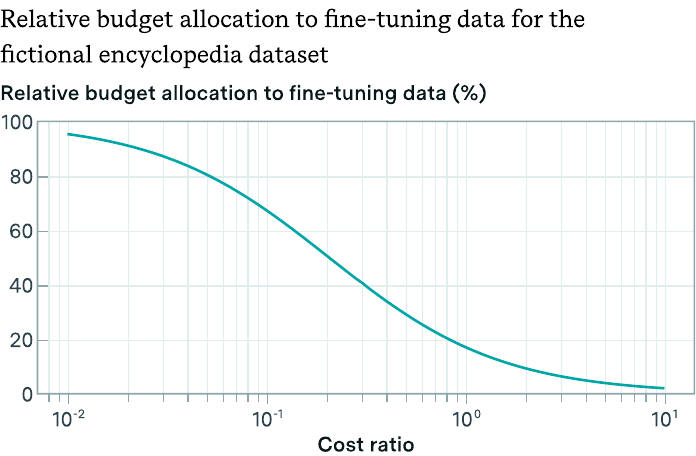}
        \caption{Budget allocation vs. cost ratio \( C_f / C_p \).}
        \label{fig:evolution_of_fine_tuning_by_cost_ratio}
    \end{subfigure}
    \caption{The plots illustrate how the optimal budget allocation between pre-training and fine-tuning, \( p \) and \( f \), evolves under different conditions. Plot (a) shows the relationship as the transfer gap from the pre-training distribution to the fine-tuning distribution \( G \) increases, where initially the budget is largely allocated towards pre-training (\( p \)), but as \( G \) rises, the allocation shifts significantly towards fine-tuning (\( f \)). Plot (b) illustrates the budget allocation for fine-tuning data (\( f \)) as the ratio of the cost per fine-tuning data point to the cost per pre-training step (\( C_f / C_p \)) increases. Initially, the allocation towards fine-tuning (\( f \)) is higher, but as the cost ratio (\( C_f / C_p \)) increases, the budget allocation for pre-training (\( p \)) also increases, reflecting the impact of cost ratio on budget optimization for a fixed transfer gap.}
    \label{fig:comparison_of_budget_allocation}
\end{figure}

If we assume the fitted scaling law found for the synthetic encyclopedia dataset, with only the transfer gap varying, Figure~\ref{fig:evolution_of_fine_tuning} illustrates how the optimal budget allocation on fine-tuning data evolves as the transfer gap \(G\) grows larger. When \(G\) is low, it is more cost-effective to allocate the budget primarily to scaling pre-training. Conversely, when \(G\) is high, it becomes more advantageous to allocate the budget predominantly to collecting fine-tuning data. This dynamic can be interpreted as indicating that pre-training is more cost-efficient when the transfer between distributions is low, underscoring the strategic importance of understanding the transfer gap when allocating training resources.

We can also consider a situation in which we fix the transfer gap, and examine the effect the relative cost between $C_f$ and $C_p$. \autoref{fig:evolution_of_fine_tuning_by_cost_ratio} illustrates how the optimal ratio of spending on fine-tuning evolves when the relative cost of fine-tuning data increases, when fixing the scaling law for transfer fit to the fictional encyclopedia dataset (including fixing the transfer gap).

Our results imply that extensive pre-training---a hallmark of the foundation model paradigm---is particularly advantageous when the transfer gap is low. When the transfer gap is low, pre-training is cost-efficient, meaning that optimally allocating the budget between pre-training and fine-tuning data to maximize downstream performance primarily involves allocating the budget towards pre-training, at least when pre-training data is cheap. 

Both \cite{mikami2021scaling} and \cite{hernandez2021scaling} found that scaling model size improved the degree of transfer. In our framework, this could potentially manifest as a decreasing transfer gap with increasing model scale, although we did not investigate the effect of model size on the transfer gap in this study. Speculatively, it is plausible that model scaling robustly decreases the transfer gap, explaining why pre-training seems to be so effective in the large model regime. We encourage future work on scaling laws for transfer to investigate this hypothesis.
\subsection{The transfer gap can set the difficulty of achieving high performance in novel domains}
On first approximation, the difficulty of training a model to perform well on a task appears to be primarily determined by the task's intrinsic complexity. However, the field of natural language processing presents a compelling counterexample. Despite the vast, inherent complexity of natural language, significant developments have been made in automating classic NLP tasks, such as translation and classification, through the development and application of large language models. We think this progress can largely be explained by the vast amount of natural language data freely available on the internet, which has enabled the training of these models on an unprecedented scale. The success in automating NLP tasks suggests that the availability of abundant, relevant training data can help overcome the challenges posed by intrinsically complex tasks.

In domains where data is less abundant, such as general-purpose robotics, we speculate that the difficulty of achieving high-performance in these domains may be significantly influenced by the transfer gap from a cheap, abundant pre-training distribution to the downstream task of interest. This is because, although collecting fine-tuning data can be expensive, it may be feasible to leverage transfer learning from much cheaper pre-training distributions with additional model scaling. Indeed, roboticists have identified the hardness of effective sim2real transfer as a key difficulty in making progress in robotics (see, for example, \cite{weng2019DR}).

Given that our framework enables the direct measurement of the transfer gap, we believe it can be a valuable tool for estimating the difficulty of achieving high performance on tasks in various domains. By quantifying the transfer gap between cheap, abundant pre-training data (such as internet text data) and the target task data, researchers can gain insights into the potential challenges and feasibility of achieving high performance on specific tasks using machine learning. This information can help guide decisions on resource allocation, such as whether to focus on collecting more task-specific data or investing in larger-scale pre-training. Ultimately, understanding the transfer gap may be key to unlocking progress in domains where data scarcity has been a significant barrier to performance.

\section{Conclusion}
In this study, we presented an empirical analysis of scaling laws for transfer learning in transformer language models. By fitting a scaling law that incorporates terms for pre-training data and fine-tuning data size, we were able to measure the degree of transfer from pre-training to various downstream distributions. This scaling law provided a good fit across several language datasets, and reveals significant variation in the magnitude of the transfer gaps.

A key positive result was that the standard errors for all parameter estimates were relatively tight, even though we used only 150 data points per fine-tuning dataset. Given that we achieved useful results without significant computing resources, we believe that our study has illustrated the potential for using scaling laws for transfer to precisely measure the transfer gap between distributions, even without expensive computing setups.

Our work has important implications for the development of foundation models and achieving higher performance on complex tasks. When the transfer gap is sufficiently small, pre-training can become a highly cost-effective strategy for enhancing downstream performance. Conversely, for tasks with large transfer gaps, focusing resources on curating fine-tuning data may be more impactful. By accurately measuring transfer gaps across a wide range of domains, we can develop a clearer picture of the relative difficulty of achieving high performance on different tasks using deep learning and guide resource allocation accordingly.

However, we also caution against interpreting these results too broadly. Since we did not examine the effect of model scale or model architecture on scaling laws for transfer, it is not straightforward to apply these scaling laws to models in completely different contexts. While we hope that this strand of research can be useful for helping researchers allocate resources between pre-training and fine-tuning to achieve high downstream performance, a more complete investigation is needed before this work can be applied directly to many applications.

\printbibliography

@misc{hernandez2021scaling,
      title={Scaling Laws for Transfer}, 
      author={Danny Hernandez and Jared Kaplan and Tom Henighan and Sam McCandlish},
      year={2021},
      eprint={2102.01293},
      archivePrefix={arXiv},
      primaryClass={cs.LG}
}

@article{hu2021lora,
  title={Lora: Low-rank adaptation of large language models},
  author={Hu, Edward J and Shen, Yelong and Wallis, Phillip and Allen-Zhu, Zeyuan and Li, Yuanzhi and Wang, Shean and Wang, Lu and Chen, Weizhu},
  journal={arXiv preprint arXiv:2106.09685},
  year={2021}
}

@misc{bommasani2022opportunities,
      title={On the Opportunities and Risks of Foundation Models}, 
      author={Rishi Bommasani and Drew A. Hudson and Ehsan Adeli and Russ Altman and Simran Arora and Sydney von Arx and Michael S. Bernstein and Jeannette Bohg and Antoine Bosselut and Emma Brunskill and Erik Brynjolfsson and Shyamal Buch and Dallas Card and Rodrigo Castellon and Niladri Chatterji and Annie Chen and Kathleen Creel and Jared Quincy Davis and Dora Demszky and Chris Donahue and Moussa Doumbouya and Esin Durmus and Stefano Ermon and John Etchemendy and Kawin Ethayarajh and Li Fei-Fei and Chelsea Finn and Trevor Gale and Lauren Gillespie and Karan Goel and Noah Goodman and Shelby Grossman and Neel Guha and Tatsunori Hashimoto and Peter Henderson and John Hewitt and Daniel E. Ho and Jenny Hong and Kyle Hsu and Jing Huang and Thomas Icard and Saahil Jain and Dan Jurafsky and Pratyusha Kalluri and Siddharth Karamcheti and Geoff Keeling and Fereshte Khani and Omar Khattab and Pang Wei Koh and Mark Krass and Ranjay Krishna and Rohith Kuditipudi and Ananya Kumar and Faisal Ladhak and Mina Lee and Tony Lee and Jure Leskovec and Isabelle Levent and Xiang Lisa Li and Xuechen Li and Tengyu Ma and Ali Malik and Christopher D. Manning and Suvir Mirchandani and Eric Mitchell and Zanele Munyikwa and Suraj Nair and Avanika Narayan and Deepak Narayanan and Ben Newman and Allen Nie and Juan Carlos Niebles and Hamed Nilforoshan and Julian Nyarko and Giray Ogut and Laurel Orr and Isabel Papadimitriou and Joon Sung Park and Chris Piech and Eva Portelance and Christopher Potts and Aditi Raghunathan and Rob Reich and Hongyu Ren and Frieda Rong and Yusuf Roohani and Camilo Ruiz and Jack Ryan and Christopher Ré and Dorsa Sadigh and Shiori Sagawa and Keshav Santhanam and Andy Shih and Krishnan Srinivasan and Alex Tamkin and Rohan Taori and Armin W. Thomas and Florian Tramèr and Rose E. Wang and William Wang and Bohan Wu and Jiajun Wu and Yuhuai Wu and Sang Michael Xie and Michihiro Yasunaga and Jiaxuan You and Matei Zaharia and Michael Zhang and Tianyi Zhang and Xikun Zhang and Yuhui Zhang and Lucia Zheng and Kaitlyn Zhou and Percy Liang},
      year={2022},
      eprint={2108.07258},
      archivePrefix={arXiv},
      primaryClass={cs.LG}
}

@misc{mikami2021scaling,
      title={A Scaling Law for Synthetic-to-Real Transfer: How Much Is Your Pre-training Effective?}, 
      author={Hiroaki Mikami and Kenji Fukumizu and Shogo Murai and Shuji Suzuki and Yuta Kikuchi and Taiji Suzuki and Shin-ichi Maeda and Kohei Hayashi},
      year={2021},
      eprint={2108.11018},
      archivePrefix={arXiv},
      primaryClass={cs.LG}
}

@misc{enron_dataset,
  title        = {The Enron Email Dataset},
  author       = {Cohen, William W.},
  howpublished = {Available for research use},
  year         = {2009},
  note         = {Provided by Carnegie Mellon University's School of Computer Science}
}

@misc{epoch2023pcdtrends,
  title="Parameter, Compute and Data Trends in Machine Learning",
  author={{Epoch AI}},
  year=2024,
  url={https://epochai.org/data/epochdb/visualization},
  note={Accessed: 2024-05-22}
}

@misc{tripuraneni2020theory,
      title={On the Theory of Transfer Learning: The Importance of Task Diversity}, 
      author={Nilesh Tripuraneni and Michael I. Jordan and Chi Jin},
      year={2020},
      eprint={2006.11650},
      archivePrefix={arXiv},
      primaryClass={cs.LG}
}

@misc{besiroglu2024chinchilla,
      title={Chinchilla Scaling: A replication attempt}, 
      author={Tamay Besiroglu and Ege Erdil and Matthew Barnett and Josh You},
      year={2024},
      eprint={2404.10102},
      archivePrefix={arXiv},
      primaryClass={cs.AI}
}

@misc{geminiteam2024gemini,
      title={Gemini: A Family of Highly Capable Multimodal Models}, 
      author={Gemini Team and Rohan Anil and Sebastian Borgeaud and Jean-Baptiste Alayrac and Jiahui Yu and Radu Soricut and Johan Schalkwyk and Andrew M. Dai and Anja Hauth and Katie Millican and David Silver and Melvin Johnson and Ioannis Antonoglou and Julian Schrittwieser and Amelia Glaese and Jilin Chen and Emily Pitler and Timothy Lillicrap and Angeliki Lazaridou and Orhan Firat and James Molloy and Michael Isard and Paul R. Barham and Tom Hennigan and Benjamin Lee and Fabio Viola and Malcolm Reynolds and Yuanzhong Xu and Ryan Doherty and Eli Collins and Clemens Meyer and Eliza Rutherford and Erica Moreira and Kareem Ayoub and Megha Goel and Jack Krawczyk and Cosmo Du and Ed Chi and Heng-Tze Cheng and Eric Ni and Purvi Shah and Patrick Kane and Betty Chan and Manaal Faruqui and Aliaksei Severyn and Hanzhao Lin and YaGuang Li and Yong Cheng and Abe Ittycheriah and Mahdis Mahdieh and Mia Chen and Pei Sun and Dustin Tran and Sumit Bagri and Balaji Lakshminarayanan and Jeremiah Liu and Andras Orban and Fabian Güra and Hao Zhou and Xinying Song and Aurelien Boffy and Harish Ganapathy and Steven Zheng and HyunJeong Choe and Ágoston Weisz and Tao Zhu and Yifeng Lu and Siddharth Gopal and Jarrod Kahn and Maciej Kula and Jeff Pitman and Rushin Shah and Emanuel Taropa and Majd Al Merey and Martin Baeuml and Zhifeng Chen and Laurent El Shafey and Yujing Zhang and Olcan Sercinoglu and George Tucker and Enrique Piqueras and Maxim Krikun and Iain Barr and Nikolay Savinov and Ivo Danihelka and Becca Roelofs and Anaïs White and Anders Andreassen and Tamara von Glehn and Lakshman Yagati and Mehran Kazemi and Lucas Gonzalez and Misha Khalman and Jakub Sygnowski and Alexandre Frechette and Charlotte Smith and Laura Culp and Lev Proleev and Yi Luan and Xi Chen and James Lottes and Nathan Schucher and Federico Lebron and Alban Rrustemi and Natalie Clay and Phil Crone and Tomas Kocisky and Jeffrey Zhao and Bartek Perz and Dian Yu and Heidi Howard and Adam Bloniarz and Jack W. Rae and Han Lu and Laurent Sifre and Marcello Maggioni and Fred Alcober and Dan Garrette and Megan Barnes and Shantanu Thakoor and Jacob Austin and Gabriel Barth-Maron and William Wong and Rishabh Joshi and Rahma Chaabouni and Deeni Fatiha and Arun Ahuja and Gaurav Singh Tomar and Evan Senter and Martin Chadwick and Ilya Kornakov and Nithya Attaluri and Iñaki Iturrate and Ruibo Liu and Yunxuan Li and Sarah Cogan and Jeremy Chen and Chao Jia and Chenjie Gu and Qiao Zhang and Jordan Grimstad and Ale Jakse Hartman and Xavier Garcia and Thanumalayan Sankaranarayana Pillai and Jacob Devlin and Michael Laskin and Diego de Las Casas and Dasha Valter and Connie Tao and Lorenzo Blanco and Adrià Puigdomènech Badia and David Reitter and Mianna Chen and Jenny Brennan and Clara Rivera and Sergey Brin and Shariq Iqbal and Gabriela Surita and Jane Labanowski and Abhi Rao and Stephanie Winkler and Emilio Parisotto and Yiming Gu and Kate Olszewska and Ravi Addanki and Antoine Miech and Annie Louis and Denis Teplyashin and Geoff Brown and Elliot Catt and Jan Balaguer and Jackie Xiang and Pidong Wang and Zoe Ashwood and Anton Briukhov and Albert Webson and Sanjay Ganapathy and Smit Sanghavi and Ajay Kannan and Ming-Wei Chang and Axel Stjerngren and Josip Djolonga and Yuting Sun and Ankur Bapna and Matthew Aitchison and Pedram Pejman and Henryk Michalewski and Tianhe Yu and Cindy Wang and Juliette Love and Junwhan Ahn and Dawn Bloxwich and Kehang Han and Peter Humphreys and Thibault Sellam and James Bradbury and Varun Godbole and Sina Samangooei and Bogdan Damoc and Alex Kaskasoli and Sébastien M. R. Arnold and Vijay Vasudevan and Shubham Agrawal and Jason Riesa and Dmitry Lepikhin and Richard Tanburn and Srivatsan Srinivasan and Hyeontaek Lim and Sarah Hodkinson and Pranav Shyam and Johan Ferret and Steven Hand and Ankush Garg and Tom Le Paine and Jian Li and Yujia Li and Minh Giang and Alexander Neitz and Zaheer Abbas and Sarah York and Machel Reid and Elizabeth Cole and Aakanksha Chowdhery and Dipanjan Das and Dominika Rogozińska and Vitaliy Nikolaev and Pablo Sprechmann and Zachary Nado and Lukas Zilka and Flavien Prost and Luheng He and Marianne Monteiro and Gaurav Mishra and Chris Welty and Josh Newlan and Dawei Jia and Miltiadis Allamanis and Clara Huiyi Hu and Raoul de Liedekerke and Justin Gilmer and Carl Saroufim and Shruti Rijhwani and Shaobo Hou and Disha Shrivastava and Anirudh Baddepudi and Alex Goldin and Adnan Ozturel and Albin Cassirer and Yunhan Xu and Daniel Sohn and Devendra Sachan and Reinald Kim Amplayo and Craig Swanson and Dessie Petrova and Shashi Narayan and Arthur Guez and Siddhartha Brahma and Jessica Landon and Miteyan Patel and Ruizhe Zhao and Kevin Villela and Luyu Wang and Wenhao Jia and Matthew Rahtz and Mai Giménez and Legg Yeung and James Keeling and Petko Georgiev and Diana Mincu and Boxi Wu and Salem Haykal and Rachel Saputro and Kiran Vodrahalli and James Qin and Zeynep Cankara and Abhanshu Sharma and Nick Fernando and Will Hawkins and Behnam Neyshabur and Solomon Kim and Adrian Hutter and Priyanka Agrawal and Alex Castro-Ros and George van den Driessche and Tao Wang and Fan Yang and Shuo-yiin Chang and Paul Komarek and Ross McIlroy and Mario Lučić and Guodong Zhang and Wael Farhan and Michael Sharman and Paul Natsev and Paul Michel and Yamini Bansal and Siyuan Qiao and Kris Cao and Siamak Shakeri and Christina Butterfield and Justin Chung and Paul Kishan Rubenstein and Shivani Agrawal and Arthur Mensch and Kedar Soparkar and Karel Lenc and Timothy Chung and Aedan Pope and Loren Maggiore and Jackie Kay and Priya Jhakra and Shibo Wang and Joshua Maynez and Mary Phuong and Taylor Tobin and Andrea Tacchetti and Maja Trebacz and Kevin Robinson and Yash Katariya and Sebastian Riedel and Paige Bailey and Kefan Xiao and Nimesh Ghelani and Lora Aroyo and Ambrose Slone and Neil Houlsby and Xuehan Xiong and Zhen Yang and Elena Gribovskaya and Jonas Adler and Mateo Wirth and Lisa Lee and Music Li and Thais Kagohara and Jay Pavagadhi and Sophie Bridgers and Anna Bortsova and Sanjay Ghemawat and Zafarali Ahmed and Tianqi Liu and Richard Powell and Vijay Bolina and Mariko Iinuma and Polina Zablotskaia and James Besley and Da-Woon Chung and Timothy Dozat and Ramona Comanescu and Xiance Si and Jeremy Greer and Guolong Su and Martin Polacek and Raphaël Lopez Kaufman and Simon Tokumine and Hexiang Hu and Elena Buchatskaya and Yingjie Miao and Mohamed Elhawaty and Aditya Siddhant and Nenad Tomasev and Jinwei Xing and Christina Greer and Helen Miller and Shereen Ashraf and Aurko Roy and Zizhao Zhang and Ada Ma and Angelos Filos and Milos Besta and Rory Blevins and Ted Klimenko and Chih-Kuan Yeh and Soravit Changpinyo and Jiaqi Mu and Oscar Chang and Mantas Pajarskas and Carrie Muir and Vered Cohen and Charline Le Lan and Krishna Haridasan and Amit Marathe and Steven Hansen and Sholto Douglas and Rajkumar Samuel and Mingqiu Wang and Sophia Austin and Chang Lan and Jiepu Jiang and Justin Chiu and Jaime Alonso Lorenzo and Lars Lowe Sjösund and Sébastien Cevey and Zach Gleicher and Thi Avrahami and Anudhyan Boral and Hansa Srinivasan and Vittorio Selo and Rhys May and Konstantinos Aisopos and Léonard Hussenot and Livio Baldini Soares and Kate Baumli and Michael B. Chang and Adrià Recasens and Ben Caine and Alexander Pritzel and Filip Pavetic and Fabio Pardo and Anita Gergely and Justin Frye and Vinay Ramasesh and Dan Horgan and Kartikeya Badola and Nora Kassner and Subhrajit Roy and Ethan Dyer and Víctor Campos Campos and Alex Tomala and Yunhao Tang and Dalia El Badawy and Elspeth White and Basil Mustafa and Oran Lang and Abhishek Jindal and Sharad Vikram and Zhitao Gong and Sergi Caelles and Ross Hemsley and Gregory Thornton and Fangxiaoyu Feng and Wojciech Stokowiec and Ce Zheng and Phoebe Thacker and Çağlar Ünlü and Zhishuai Zhang and Mohammad Saleh and James Svensson and Max Bileschi and Piyush Patil and Ankesh Anand and Roman Ring and Katerina Tsihlas and Arpi Vezer and Marco Selvi and Toby Shevlane and Mikel Rodriguez and Tom Kwiatkowski and Samira Daruki and Keran Rong and Allan Dafoe and Nicholas FitzGerald and Keren Gu-Lemberg and Mina Khan and Lisa Anne Hendricks and Marie Pellat and Vladimir Feinberg and James Cobon-Kerr and Tara Sainath and Maribeth Rauh and Sayed Hadi Hashemi and Richard Ives and Yana Hasson and Eric Noland and Yuan Cao and Nathan Byrd and Le Hou and Qingze Wang and Thibault Sottiaux and Michela Paganini and Jean-Baptiste Lespiau and Alexandre Moufarek and Samer Hassan and Kaushik Shivakumar and Joost van Amersfoort and Amol Mandhane and Pratik Joshi and Anirudh Goyal and Matthew Tung and Andrew Brock and Hannah Sheahan and Vedant Misra and Cheng Li and Nemanja Rakićević and Mostafa Dehghani and Fangyu Liu and Sid Mittal and Junhyuk Oh and Seb Noury and Eren Sezener and Fantine Huot and Matthew Lamm and Nicola De Cao and Charlie Chen and Sidharth Mudgal and Romina Stella and Kevin Brooks and Gautam Vasudevan and Chenxi Liu and Mainak Chain and Nivedita Melinkeri and Aaron Cohen and Venus Wang and Kristie Seymore and Sergey Zubkov and Rahul Goel and Summer Yue and Sai Krishnakumaran and Brian Albert and Nate Hurley and Motoki Sano and Anhad Mohananey and Jonah Joughin and Egor Filonov and Tomasz Kępa and Yomna Eldawy and Jiawern Lim and Rahul Rishi and Shirin Badiezadegan and Taylor Bos and Jerry Chang and Sanil Jain and Sri Gayatri Sundara Padmanabhan and Subha Puttagunta and Kalpesh Krishna and Leslie Baker and Norbert Kalb and Vamsi Bedapudi and Adam Kurzrok and Shuntong Lei and Anthony Yu and Oren Litvin and Xiang Zhou and Zhichun Wu and Sam Sobell and Andrea Siciliano and Alan Papir and Robby Neale and Jonas Bragagnolo and Tej Toor and Tina Chen and Valentin Anklin and Feiran Wang and Richie Feng and Milad Gholami and Kevin Ling and Lijuan Liu and Jules Walter and Hamid Moghaddam and Arun Kishore and Jakub Adamek and Tyler Mercado and Jonathan Mallinson and Siddhinita Wandekar and Stephen Cagle and Eran Ofek and Guillermo Garrido and Clemens Lombriser and Maksim Mukha and Botu Sun and Hafeezul Rahman Mohammad and Josip Matak and Yadi Qian and Vikas Peswani and Pawel Janus and Quan Yuan and Leif Schelin and Oana David and Ankur Garg and Yifan He and Oleksii Duzhyi and Anton Älgmyr and Timothée Lottaz and Qi Li and Vikas Yadav and Luyao Xu and Alex Chinien and Rakesh Shivanna and Aleksandr Chuklin and Josie Li and Carrie Spadine and Travis Wolfe and Kareem Mohamed and Subhabrata Das and Zihang Dai and Kyle He and Daniel von Dincklage and Shyam Upadhyay and Akanksha Maurya and Luyan Chi and Sebastian Krause and Khalid Salama and Pam G Rabinovitch and Pavan Kumar Reddy M and Aarush Selvan and Mikhail Dektiarev and Golnaz Ghiasi and Erdem Guven and Himanshu Gupta and Boyi Liu and Deepak Sharma and Idan Heimlich Shtacher and Shachi Paul and Oscar Akerlund and François-Xavier Aubet and Terry Huang and Chen Zhu and Eric Zhu and Elico Teixeira and Matthew Fritze and Francesco Bertolini and Liana-Eleonora Marinescu and Martin Bölle and Dominik Paulus and Khyatti Gupta and Tejasi Latkar and Max Chang and Jason Sanders and Roopa Wilson and Xuewei Wu and Yi-Xuan Tan and Lam Nguyen Thiet and Tulsee Doshi and Sid Lall and Swaroop Mishra and Wanming Chen and Thang Luong and Seth Benjamin and Jasmine Lee and Ewa Andrejczuk and Dominik Rabiej and Vipul Ranjan and Krzysztof Styrc and Pengcheng Yin and Jon Simon and Malcolm Rose Harriott and Mudit Bansal and Alexei Robsky and Geoff Bacon and David Greene and Daniil Mirylenka and Chen Zhou and Obaid Sarvana and Abhimanyu Goyal and Samuel Andermatt and Patrick Siegler and Ben Horn and Assaf Israel and Francesco Pongetti and Chih-Wei "Louis" Chen and Marco Selvatici and Pedro Silva and Kathie Wang and Jackson Tolins and Kelvin Guu and Roey Yogev and Xiaochen Cai and Alessandro Agostini and Maulik Shah and Hung Nguyen and Noah Ó Donnaile and Sébastien Pereira and Linda Friso and Adam Stambler and Adam Kurzrok and Chenkai Kuang and Yan Romanikhin and Mark Geller and ZJ Yan and Kane Jang and Cheng-Chun Lee and Wojciech Fica and Eric Malmi and Qijun Tan and Dan Banica and Daniel Balle and Ryan Pham and Yanping Huang and Diana Avram and Hongzhi Shi and Jasjot Singh and Chris Hidey and Niharika Ahuja and Pranab Saxena and Dan Dooley and Srividya Pranavi Potharaju and Eileen O'Neill and Anand Gokulchandran and Ryan Foley and Kai Zhao and Mike Dusenberry and Yuan Liu and Pulkit Mehta and Ragha Kotikalapudi and Chalence Safranek-Shrader and Andrew Goodman and Joshua Kessinger and Eran Globen and Prateek Kolhar and Chris Gorgolewski and Ali Ibrahim and Yang Song and Ali Eichenbaum and Thomas Brovelli and Sahitya Potluri and Preethi Lahoti and Cip Baetu and Ali Ghorbani and Charles Chen and Andy Crawford and Shalini Pal and Mukund Sridhar and Petru Gurita and Asier Mujika and Igor Petrovski and Pierre-Louis Cedoz and Chenmei Li and Shiyuan Chen and Niccolò Dal Santo and Siddharth Goyal and Jitesh Punjabi and Karthik Kappaganthu and Chester Kwak and Pallavi LV and Sarmishta Velury and Himadri Choudhury and Jamie Hall and Premal Shah and Ricardo Figueira and Matt Thomas and Minjie Lu and Ting Zhou and Chintu Kumar and Thomas Jurdi and Sharat Chikkerur and Yenai Ma and Adams Yu and Soo Kwak and Victor Ähdel and Sujeevan Rajayogam and Travis Choma and Fei Liu and Aditya Barua and Colin Ji and Ji Ho Park and Vincent Hellendoorn and Alex Bailey and Taylan Bilal and Huanjie Zhou and Mehrdad Khatir and Charles Sutton and Wojciech Rzadkowski and Fiona Macintosh and Konstantin Shagin and Paul Medina and Chen Liang and Jinjing Zhou and Pararth Shah and Yingying Bi and Attila Dankovics and Shipra Banga and Sabine Lehmann and Marissa Bredesen and Zifan Lin and John Eric Hoffmann and Jonathan Lai and Raynald Chung and Kai Yang and Nihal Balani and Arthur Bražinskas and Andrei Sozanschi and Matthew Hayes and Héctor Fernández Alcalde and Peter Makarov and Will Chen and Antonio Stella and Liselotte Snijders and Michael Mandl and Ante Kärrman and Paweł Nowak and Xinyi Wu and Alex Dyck and Krishnan Vaidyanathan and Raghavender R and Jessica Mallet and Mitch Rudominer and Eric Johnston and Sushil Mittal and Akhil Udathu and Janara Christensen and Vishal Verma and Zach Irving and Andreas Santucci and Gamaleldin Elsayed and Elnaz Davoodi and Marin Georgiev and Ian Tenney and Nan Hua and Geoffrey Cideron and Edouard Leurent and Mahmoud Alnahlawi and Ionut Georgescu and Nan Wei and Ivy Zheng and Dylan Scandinaro and Heinrich Jiang and Jasper Snoek and Mukund Sundararajan and Xuezhi Wang and Zack Ontiveros and Itay Karo and Jeremy Cole and Vinu Rajashekhar and Lara Tumeh and Eyal Ben-David and Rishub Jain and Jonathan Uesato and Romina Datta and Oskar Bunyan and Shimu Wu and John Zhang and Piotr Stanczyk and Ye Zhang and David Steiner and Subhajit Naskar and Michael Azzam and Matthew Johnson and Adam Paszke and Chung-Cheng Chiu and Jaume Sanchez Elias and Afroz Mohiuddin and Faizan Muhammad and Jin Miao and Andrew Lee and Nino Vieillard and Jane Park and Jiageng Zhang and Jeff Stanway and Drew Garmon and Abhijit Karmarkar and Zhe Dong and Jong Lee and Aviral Kumar and Luowei Zhou and Jonathan Evens and William Isaac and Geoffrey Irving and Edward Loper and Michael Fink and Isha Arkatkar and Nanxin Chen and Izhak Shafran and Ivan Petrychenko and Zhe Chen and Johnson Jia and Anselm Levskaya and Zhenkai Zhu and Peter Grabowski and Yu Mao and Alberto Magni and Kaisheng Yao and Javier Snaider and Norman Casagrande and Evan Palmer and Paul Suganthan and Alfonso Castaño and Irene Giannoumis and Wooyeol Kim and Mikołaj Rybiński and Ashwin Sreevatsa and Jennifer Prendki and David Soergel and Adrian Goedeckemeyer and Willi Gierke and Mohsen Jafari and Meenu Gaba and Jeremy Wiesner and Diana Gage Wright and Yawen Wei and Harsha Vashisht and Yana Kulizhskaya and Jay Hoover and Maigo Le and Lu Li and Chimezie Iwuanyanwu and Lu Liu and Kevin Ramirez and Andrey Khorlin and Albert Cui and Tian LIN and Marcus Wu and Ricardo Aguilar and Keith Pallo and Abhishek Chakladar and Ginger Perng and Elena Allica Abellan and Mingyang Zhang and Ishita Dasgupta and Nate Kushman and Ivo Penchev and Alena Repina and Xihui Wu and Tom van der Weide and Priya Ponnapalli and Caroline Kaplan and Jiri Simsa and Shuangfeng Li and Olivier Dousse and Fan Yang and Jeff Piper and Nathan Ie and Rama Pasumarthi and Nathan Lintz and Anitha Vijayakumar and Daniel Andor and Pedro Valenzuela and Minnie Lui and Cosmin Paduraru and Daiyi Peng and Katherine Lee and Shuyuan Zhang and Somer Greene and Duc Dung Nguyen and Paula Kurylowicz and Cassidy Hardin and Lucas Dixon and Lili Janzer and Kiam Choo and Ziqiang Feng and Biao Zhang and Achintya Singhal and Dayou Du and Dan McKinnon and Natasha Antropova and Tolga Bolukbasi and Orgad Keller and David Reid and Daniel Finchelstein and Maria Abi Raad and Remi Crocker and Peter Hawkins and Robert Dadashi and Colin Gaffney and Ken Franko and Anna Bulanova and Rémi Leblond and Shirley Chung and Harry Askham and Luis C. Cobo and Kelvin Xu and Felix Fischer and Jun Xu and Christina Sorokin and Chris Alberti and Chu-Cheng Lin and Colin Evans and Alek Dimitriev and Hannah Forbes and Dylan Banarse and Zora Tung and Mark Omernick and Colton Bishop and Rachel Sterneck and Rohan Jain and Jiawei Xia and Ehsan Amid and Francesco Piccinno and Xingyu Wang and Praseem Banzal and Daniel J. Mankowitz and Alex Polozov and Victoria Krakovna and Sasha Brown and MohammadHossein Bateni and Dennis Duan and Vlad Firoiu and Meghana Thotakuri and Tom Natan and Matthieu Geist and Ser tan Girgin and Hui Li and Jiayu Ye and Ofir Roval and Reiko Tojo and Michael Kwong and James Lee-Thorp and Christopher Yew and Danila Sinopalnikov and Sabela Ramos and John Mellor and Abhishek Sharma and Kathy Wu and David Miller and Nicolas Sonnerat and Denis Vnukov and Rory Greig and Jennifer Beattie and Emily Caveness and Libin Bai and Julian Eisenschlos and Alex Korchemniy and Tomy Tsai and Mimi Jasarevic and Weize Kong and Phuong Dao and Zeyu Zheng and Frederick Liu and Fan Yang and Rui Zhu and Tian Huey Teh and Jason Sanmiya and Evgeny Gladchenko and Nejc Trdin and Daniel Toyama and Evan Rosen and Sasan Tavakkol and Linting Xue and Chen Elkind and Oliver Woodman and John Carpenter and George Papamakarios and Rupert Kemp and Sushant Kafle and Tanya Grunina and Rishika Sinha and Alice Talbert and Diane Wu and Denese Owusu-Afriyie and Cosmo Du and Chloe Thornton and Jordi Pont-Tuset and Pradyumna Narayana and Jing Li and Saaber Fatehi and John Wieting and Omar Ajmeri and Benigno Uria and Yeongil Ko and Laura Knight and Amélie Héliou and Ning Niu and Shane Gu and Chenxi Pang and Yeqing Li and Nir Levine and Ariel Stolovich and Rebeca Santamaria-Fernandez and Sonam Goenka and Wenny Yustalim and Robin Strudel and Ali Elqursh and Charlie Deck and Hyo Lee and Zonglin Li and Kyle Levin and Raphael Hoffmann and Dan Holtmann-Rice and Olivier Bachem and Sho Arora and Christy Koh and Soheil Hassas Yeganeh and Siim Põder and Mukarram Tariq and Yanhua Sun and Lucian Ionita and Mojtaba Seyedhosseini and Pouya Tafti and Zhiyu Liu and Anmol Gulati and Jasmine Liu and Xinyu Ye and Bart Chrzaszcz and Lily Wang and Nikhil Sethi and Tianrun Li and Ben Brown and Shreya Singh and Wei Fan and Aaron Parisi and Joe Stanton and Vinod Koverkathu and Christopher A. Choquette-Choo and Yunjie Li and TJ Lu and Abe Ittycheriah and Prakash Shroff and Mani Varadarajan and Sanaz Bahargam and Rob Willoughby and David Gaddy and Guillaume Desjardins and Marco Cornero and Brona Robenek and Bhavishya Mittal and Ben Albrecht and Ashish Shenoy and Fedor Moiseev and Henrik Jacobsson and Alireza Ghaffarkhah and Morgane Rivière and Alanna Walton and Clément Crepy and Alicia Parrish and Zongwei Zhou and Clement Farabet and Carey Radebaugh and Praveen Srinivasan and Claudia van der Salm and Andreas Fidjeland and Salvatore Scellato and Eri Latorre-Chimoto and Hanna Klimczak-Plucińska and David Bridson and Dario de Cesare and Tom Hudson and Piermaria Mendolicchio and Lexi Walker and Alex Morris and Matthew Mauger and Alexey Guseynov and Alison Reid and Seth Odoom and Lucia Loher and Victor Cotruta and Madhavi Yenugula and Dominik Grewe and Anastasia Petrushkina and Tom Duerig and Antonio Sanchez and Steve Yadlowsky and Amy Shen and Amir Globerson and Lynette Webb and Sahil Dua and Dong Li and Surya Bhupatiraju and Dan Hurt and Haroon Qureshi and Ananth Agarwal and Tomer Shani and Matan Eyal and Anuj Khare and Shreyas Rammohan Belle and Lei Wang and Chetan Tekur and Mihir Sanjay Kale and Jinliang Wei and Ruoxin Sang and Brennan Saeta and Tyler Liechty and Yi Sun and Yao Zhao and Stephan Lee and Pandu Nayak and Doug Fritz and Manish Reddy Vuyyuru and John Aslanides and Nidhi Vyas and Martin Wicke and Xiao Ma and Evgenii Eltyshev and Nina Martin and Hardie Cate and James Manyika and Keyvan Amiri and Yelin Kim and Xi Xiong and Kai Kang and Florian Luisier and Nilesh Tripuraneni and David Madras and Mandy Guo and Austin Waters and Oliver Wang and Joshua Ainslie and Jason Baldridge and Han Zhang and Garima Pruthi and Jakob Bauer and Feng Yang and Riham Mansour and Jason Gelman and Yang Xu and George Polovets and Ji Liu and Honglong Cai and Warren Chen and XiangHai Sheng and Emily Xue and Sherjil Ozair and Christof Angermueller and Xiaowei Li and Anoop Sinha and Weiren Wang and Julia Wiesinger and Emmanouil Koukoumidis and Yuan Tian and Anand Iyer and Madhu Gurumurthy and Mark Goldenson and Parashar Shah and MK Blake and Hongkun Yu and Anthony Urbanowicz and Jennimaria Palomaki and Chrisantha Fernando and Ken Durden and Harsh Mehta and Nikola Momchev and Elahe Rahimtoroghi and Maria Georgaki and Amit Raul and Sebastian Ruder and Morgan Redshaw and Jinhyuk Lee and Denny Zhou and Komal Jalan and Dinghua Li and Blake Hechtman and Parker Schuh and Milad Nasr and Kieran Milan and Vladimir Mikulik and Juliana Franco and Tim Green and Nam Nguyen and Joe Kelley and Aroma Mahendru and Andrea Hu and Joshua Howland and Ben Vargas and Jeffrey Hui and Kshitij Bansal and Vikram Rao and Rakesh Ghiya and Emma Wang and Ke Ye and Jean Michel Sarr and Melanie Moranski Preston and Madeleine Elish and Steve Li and Aakash Kaku and Jigar Gupta and Ice Pasupat and Da-Cheng Juan and Milan Someswar and Tejvi M. and Xinyun Chen and Aida Amini and Alex Fabrikant and Eric Chu and Xuanyi Dong and Amruta Muthal and Senaka Buthpitiya and Sarthak Jauhari and Nan Hua and Urvashi Khandelwal and Ayal Hitron and Jie Ren and Larissa Rinaldi and Shahar Drath and Avigail Dabush and Nan-Jiang Jiang and Harshal Godhia and Uli Sachs and Anthony Chen and Yicheng Fan and Hagai Taitelbaum and Hila Noga and Zhuyun Dai and James Wang and Chen Liang and Jenny Hamer and Chun-Sung Ferng and Chenel Elkind and Aviel Atias and Paulina Lee and Vít Listík and Mathias Carlen and Jan van de Kerkhof and Marcin Pikus and Krunoslav Zaher and Paul Müller and Sasha Zykova and Richard Stefanec and Vitaly Gatsko and Christoph Hirnschall and Ashwin Sethi and Xingyu Federico Xu and Chetan Ahuja and Beth Tsai and Anca Stefanoiu and Bo Feng and Keshav Dhandhania and Manish Katyal and Akshay Gupta and Atharva Parulekar and Divya Pitta and Jing Zhao and Vivaan Bhatia and Yashodha Bhavnani and Omar Alhadlaq and Xiaolin Li and Peter Danenberg and Dennis Tu and Alex Pine and Vera Filippova and Abhipso Ghosh and Ben Limonchik and Bhargava Urala and Chaitanya Krishna Lanka and Derik Clive and Yi Sun and Edward Li and Hao Wu and Kevin Hongtongsak and Ianna Li and Kalind Thakkar and Kuanysh Omarov and Kushal Majmundar and Michael Alverson and Michael Kucharski and Mohak Patel and Mudit Jain and Maksim Zabelin and Paolo Pelagatti and Rohan Kohli and Saurabh Kumar and Joseph Kim and Swetha Sankar and Vineet Shah and Lakshmi Ramachandruni and Xiangkai Zeng and Ben Bariach and Laura Weidinger and Amar Subramanya and Sissie Hsiao and Demis Hassabis and Koray Kavukcuoglu and Adam Sadovsky and Quoc Le and Trevor Strohman and Yonghui Wu and Slav Petrov and Jeffrey Dean and Oriol Vinyals},
      year={2024},
      eprint={2312.11805},
      archivePrefix={arXiv},
      primaryClass={cs.CL}
}

@misc{mosbach2023fewshot,
      title={Few-shot Fine-tuning vs. In-context Learning: A Fair Comparison and Evaluation}, 
      author={Marius Mosbach and Tiago Pimentel and Shauli Ravfogel and Dietrich Klakow and Yanai Elazar},
      year={2023},
      eprint={2305.16938},
      archivePrefix={arXiv},
      primaryClass={cs.CL}
}

@misc{henighan2020scaling,
      title={Scaling Laws for Autoregressive Generative Modeling}, 
      author={Tom Henighan and Jared Kaplan and Mor Katz and Mark Chen and Christopher Hesse and Jacob Jackson and Heewoo Jun and Tom B. Brown and Prafulla Dhariwal and Scott Gray and Chris Hallacy and Benjamin Mann and Alec Radford and Aditya Ramesh and Nick Ryder and Daniel M. Ziegler and John Schulman and Dario Amodei and Sam McCandlish},
      year={2020},
      eprint={2010.14701},
      archivePrefix={arXiv},
      primaryClass={cs.LG}
}

@misc{maurer2016benefit,
      title={The Benefit of Multitask Representation Learning}, 
      author={Andreas Maurer and Massimiliano Pontil and Bernardino Romera-Paredes},
      year={2016},
      eprint={1505.06279},
      archivePrefix={arXiv},
      primaryClass={stat.ML}
}

@article{Baxter_2000,
   title={A Model of Inductive Bias Learning},
   volume={12},
   ISSN={1076-9757},
   url={http://dx.doi.org/10.1613/jair.731},
   DOI={10.1613/jair.731},
   journal={Journal of Artificial Intelligence Research},
   publisher={AI Access Foundation},
   author={Baxter, J.},
   year={2000},
   month=mar, pages={149–198} }

@misc{zhang2024scaling,
      title={When Scaling Meets LLM Finetuning: The Effect of Data, Model and Finetuning Method}, 
      author={Biao Zhang and Zhongtao Liu and Colin Cherry and Orhan Firat},
      year={2024},
      eprint={2402.17193},
      archivePrefix={arXiv},
      primaryClass={cs.CL}
}

@article{weng2019DR,
  title   = "Domain Randomization for Sim2Real Transfer",
  author  = "Weng, Lilian",
  journal = "lilianweng.github.io",
  year    = "2019",
  url     = "https://lilianweng.github.io/posts/2019-05-05-domain-randomization/"
}

@misc{biderman2023pythia,
      title={Pythia: A Suite for Analyzing Large Language Models Across Training and Scaling}, 
      author={Stella Biderman and Hailey Schoelkopf and Quentin Anthony and Herbie Bradley and Kyle O'Brien and Eric Hallahan and Mohammad Aflah Khan and Shivanshu Purohit and USVSN Sai Prashanth and Edward Raff and Aviya Skowron and Lintang Sutawika and Oskar van der Wal},
      year={2023},
      eprint={2304.01373},
      archivePrefix={arXiv},
      primaryClass={cs.CL}
}

@misc{epoch2023scalinglawsliteraturereview,
  title = "Scaling Laws Literature Review",
  author = {Pablo Villalobos},
  year = 2023,
  url = {https://epochai.org/blog/scaling-laws-literature-review},
  note = "Accessed: 2023-9-12"
}

@misc{hestness2017deep,
      title={Deep Learning Scaling is Predictable, Empirically}, 
      author={Joel Hestness and Sharan Narang and Newsha Ardalani and Gregory Diamos and Heewoo Jun and Hassan Kianinejad and Md. Mostofa Ali Patwary and Yang Yang and Yanqi Zhou},
      year={2017},
      eprint={1712.00409},
      archivePrefix={arXiv},
      primaryClass={cs.LG}
}

@misc{rosenfeld2019constructive,
      title={A Constructive Prediction of the Generalization Error Across Scales}, 
      author={Jonathan S. Rosenfeld and Amir Rosenfeld and Yonatan Belinkov and Nir Shavit},
      year={2019},
      eprint={1909.12673},
      archivePrefix={arXiv},
      primaryClass={cs.LG}
}

@misc{kaplan2020scaling,
      title={Scaling Laws for Neural Language Models}, 
      author={Jared Kaplan and Sam McCandlish and Tom Henighan and Tom B. Brown and Benjamin Chess and Rewon Child and Scott Gray and Alec Radford and Jeffrey Wu and Dario Amodei},
      year={2020},
      eprint={2001.08361},
      archivePrefix={arXiv},
      primaryClass={cs.LG}
}

@misc{hoffmann2022training,
      title={Training Compute-Optimal Large Language Models}, 
      author={Jordan Hoffmann and Sebastian Borgeaud and Arthur Mensch and Elena Buchatskaya and Trevor Cai and Eliza Rutherford and Diego de Las Casas and Lisa Anne Hendricks and Johannes Welbl and Aidan Clark and Tom Hennigan and Eric Noland and Katie Millican and George van den Driessche and Bogdan Damoc and Aurelia Guy and Simon Osindero and Karen Simonyan and Erich Elsen and Jack W. Rae and Oriol Vinyals and Laurent Sifre},
      year={2022},
      eprint={2203.15556},
      archivePrefix={arXiv},
      primaryClass={cs.CL}
}

@misc{arora2019finegrained,
      title={Fine-Grained Analysis of Optimization and Generalization for Overparameterized Two-Layer Neural Networks}, 
      author={Sanjeev Arora and Simon S. Du and Wei Hu and Zhiyuan Li and Ruosong Wang},
      year={2019},
      eprint={1901.08584},
      archivePrefix={arXiv},
      primaryClass={cs.LG}
}

@misc{nitanda2020gradient,
      title={Gradient Descent can Learn Less Over-parameterized Two-layer Neural Networks on Classification Problems}, 
      author={Atsushi Nitanda and Geoffrey Chinot and Taiji Suzuki},
      year={2020},
      eprint={1905.09870},
      archivePrefix={arXiv},
      primaryClass={stat.ML}
}

@misc{gao2020pile,
      title={The Pile: An 800GB Dataset of Diverse Text for Language Modeling}, 
      author={Leo Gao and Stella Biderman and Sid Black and Laurence Golding and Travis Hoppe and Charles Foster and Jason Phang and Horace He and Anish Thite and Noa Nabeshima and Shawn Presser and Connor Leahy},
      year={2020},
      eprint={2101.00027},
      archivePrefix={arXiv},
      primaryClass={cs.CL}
}

@misc{fieberg2024statistics,
  author = {Fieberg, John R.},
  title = {Statistics for Ecologists: A Frequentist and Bayesian Treatment of Modern Regression Models},
  year = {2024},
  publisher = {University of Minnesota Libraries Publishing},
  howpublished = {Retrieved from the University of Minnesota Digital Conservancy},
  url = {https://hdl.handle.net/11299/260227}
}

@misc{felis_catus_9_ensemble,
  author = {{Genome Sequencing Center at Washington University School of Medicine}},
  title = {Cat Assembly and Gene Annotation},
  year = {2020},
  month = {September},
  note = {Assembly: Felis\_catus\_9.0, INSDC Assembly GCA\_000181335.4, Nov 2017. Genebuild last updated/patched September 2020. Database version 111.9. Ensembl release 111 - January 2024.},
  url = {http://www.ensembl.org/Felis_catus/Info/Index}
}

@misc{raffel2023exploring,
      title={Exploring the Limits of Transfer Learning with a Unified Text-to-Text Transformer}, 
      author={Colin Raffel and Noam Shazeer and Adam Roberts and Katherine Lee and Sharan Narang and Michael Matena and Yanqi Zhou and Wei Li and Peter J. Liu},
      year={2023},
      eprint={1910.10683},
      archivePrefix={arXiv},
      primaryClass={cs.LG}
}

@misc{jiang2024mixtral,
      title={Mixtral of Experts}, 
      author={Albert Q. Jiang and Alexandre Sablayrolles and Antoine Roux and Arthur Mensch and Blanche Savary and Chris Bamford and Devendra Singh Chaplot and Diego de las Casas and Emma Bou Hanna and Florian Bressand and Gianna Lengyel and Guillaume Bour and Guillaume Lample and Lélio Renard Lavaud and Lucile Saulnier and Marie-Anne Lachaux and Pierre Stock and Sandeep Subramanian and Sophia Yang and Szymon Antoniak and Teven Le Scao and Théophile Gervet and Thibaut Lavril and Thomas Wang and Timothée Lacroix and William El Sayed},
      year={2024},
      eprint={2401.04088},
      archivePrefix={arXiv},
      primaryClass={cs.LG}
}

@misc{geminiteam2023gemini,
      title={Gemini: A Family of Highly Capable Multimodal Models}, 
      author={Gemini Team and Rohan Anil and Sebastian Borgeaud and Yonghui Wu and Jean-Baptiste Alayrac and Jiahui Yu and Radu Soricut and Johan Schalkwyk and Andrew M. Dai and Anja Hauth and Katie Millican and David Silver and Slav Petrov and Melvin Johnson and Ioannis Antonoglou and Julian Schrittwieser and Amelia Glaese and Jilin Chen and Emily Pitler and Timothy Lillicrap and Angeliki Lazaridou and Orhan Firat and James Molloy and Michael Isard and Paul R. Barham and Tom Hennigan and Benjamin Lee and Fabio Viola and Malcolm Reynolds and Yuanzhong Xu and Ryan Doherty and Eli Collins and Clemens Meyer and Eliza Rutherford and Erica Moreira and Kareem Ayoub and Megha Goel and George Tucker and Enrique Piqueras and Maxim Krikun and Iain Barr and Nikolay Savinov and Ivo Danihelka and Becca Roelofs and Anaïs White and Anders Andreassen and Tamara von Glehn and Lakshman Yagati and Mehran Kazemi and Lucas Gonzalez and Misha Khalman and Jakub Sygnowski and Alexandre Frechette and Charlotte Smith and Laura Culp and Lev Proleev and Yi Luan and Xi Chen and James Lottes and Nathan Schucher and Federico Lebron and Alban Rrustemi and Natalie Clay and Phil Crone and Tomas Kocisky and Jeffrey Zhao and Bartek Perz and Dian Yu and Heidi Howard and Adam Bloniarz and Jack W. Rae and Han Lu and Laurent Sifre and Marcello Maggioni and Fred Alcober and Dan Garrette and Megan Barnes and Shantanu Thakoor and Jacob Austin and Gabriel Barth-Maron and William Wong and Rishabh Joshi and Rahma Chaabouni and Deeni Fatiha and Arun Ahuja and Ruibo Liu and Yunxuan Li and Sarah Cogan and Jeremy Chen and Chao Jia and Chenjie Gu and Qiao Zhang and Jordan Grimstad and Ale Jakse Hartman and Martin Chadwick and Gaurav Singh Tomar and Xavier Garcia and Evan Senter and Emanuel Taropa and Thanumalayan Sankaranarayana Pillai and Jacob Devlin and Michael Laskin and Diego de Las Casas and Dasha Valter and Connie Tao and Lorenzo Blanco and Adrià Puigdomènech Badia and David Reitter and Mianna Chen and Jenny Brennan and Clara Rivera and Sergey Brin and Shariq Iqbal and Gabriela Surita and Jane Labanowski and Abhi Rao and Stephanie Winkler and Emilio Parisotto and Yiming Gu and Kate Olszewska and Yujing Zhang and Ravi Addanki and Antoine Miech and Annie Louis and Laurent El Shafey and Denis Teplyashin and Geoff Brown and Elliot Catt and Nithya Attaluri and Jan Balaguer and Jackie Xiang and Pidong Wang and Zoe Ashwood and Anton Briukhov and Albert Webson and Sanjay Ganapathy and Smit Sanghavi and Ajay Kannan and Ming-Wei Chang and Axel Stjerngren and Josip Djolonga and Yuting Sun and Ankur Bapna and Matthew Aitchison and Pedram Pejman and Henryk Michalewski and Tianhe Yu and Cindy Wang and Juliette Love and Junwhan Ahn and Dawn Bloxwich and Kehang Han and Peter Humphreys and Thibault Sellam and James Bradbury and Varun Godbole and Sina Samangooei and Bogdan Damoc and Alex Kaskasoli and Sébastien M. R. Arnold and Vijay Vasudevan and Shubham Agrawal and Jason Riesa and Dmitry Lepikhin and Richard Tanburn and Srivatsan Srinivasan and Hyeontaek Lim and Sarah Hodkinson and Pranav Shyam and Johan Ferret and Steven Hand and Ankush Garg and Tom Le Paine and Jian Li and Yujia Li and Minh Giang and Alexander Neitz and Zaheer Abbas and Sarah York and Machel Reid and Elizabeth Cole and Aakanksha Chowdhery and Dipanjan Das and Dominika Rogozińska and Vitaly Nikolaev and Pablo Sprechmann and Zachary Nado and Lukas Zilka and Flavien Prost and Luheng He and Marianne Monteiro and Gaurav Mishra and Chris Welty and Josh Newlan and Dawei Jia and Miltiadis Allamanis and Clara Huiyi Hu and Raoul de Liedekerke and Justin Gilmer and Carl Saroufim and Shruti Rijhwani and Shaobo Hou and Disha Shrivastava and Anirudh Baddepudi and Alex Goldin and Adnan Ozturel and Albin Cassirer and Yunhan Xu and Daniel Sohn and Devendra Sachan and Reinald Kim Amplayo and Craig Swanson and Dessie Petrova and Shashi Narayan and Arthur Guez and Siddhartha Brahma and Jessica Landon and Miteyan Patel and Ruizhe Zhao and Kevin Villela and Luyu Wang and Wenhao Jia and Matthew Rahtz and Mai Giménez and Legg Yeung and Hanzhao Lin and James Keeling and Petko Georgiev and Diana Mincu and Boxi Wu and Salem Haykal and Rachel Saputro and Kiran Vodrahalli and James Qin and Zeynep Cankara and Abhanshu Sharma and Nick Fernando and Will Hawkins and Behnam Neyshabur and Solomon Kim and Adrian Hutter and Priyanka Agrawal and Alex Castro-Ros and George van den Driessche and Tao Wang and Fan Yang and Shuo-yiin Chang and Paul Komarek and Ross McIlroy and Mario Lučić and Guodong Zhang and Wael Farhan and Michael Sharman and Paul Natsev and Paul Michel and Yong Cheng and Yamini Bansal and Siyuan Qiao and Kris Cao and Siamak Shakeri and Christina Butterfield and Justin Chung and Paul Kishan Rubenstein and Shivani Agrawal and Arthur Mensch and Kedar Soparkar and Karel Lenc and Timothy Chung and Aedan Pope and Loren Maggiore and Jackie Kay and Priya Jhakra and Shibo Wang and Joshua Maynez and Mary Phuong and Taylor Tobin and Andrea Tacchetti and Maja Trebacz and Kevin Robinson and Yash Katariya and Sebastian Riedel and Paige Bailey and Kefan Xiao and Nimesh Ghelani and Lora Aroyo and Ambrose Slone and Neil Houlsby and Xuehan Xiong and Zhen Yang and Elena Gribovskaya and Jonas Adler and Mateo Wirth and Lisa Lee and Music Li and Thais Kagohara and Jay Pavagadhi and Sophie Bridgers and Anna Bortsova and Sanjay Ghemawat and Zafarali Ahmed and Tianqi Liu and Richard Powell and Vijay Bolina and Mariko Iinuma and Polina Zablotskaia and James Besley and Da-Woon Chung and Timothy Dozat and Ramona Comanescu and Xiance Si and Jeremy Greer and Guolong Su and Martin Polacek and Raphaël Lopez Kaufman and Simon Tokumine and Hexiang Hu and Elena Buchatskaya and Yingjie Miao and Mohamed Elhawaty and Aditya Siddhant and Nenad Tomasev and Jinwei Xing and Christina Greer and Helen Miller and Shereen Ashraf and Aurko Roy and Zizhao Zhang and Ada Ma and Angelos Filos and Milos Besta and Rory Blevins and Ted Klimenko and Chih-Kuan Yeh and Soravit Changpinyo and Jiaqi Mu and Oscar Chang and Mantas Pajarskas and Carrie Muir and Vered Cohen and Charline Le Lan and Krishna Haridasan and Amit Marathe and Steven Hansen and Sholto Douglas and Rajkumar Samuel and Mingqiu Wang and Sophia Austin and Chang Lan and Jiepu Jiang and Justin Chiu and Jaime Alonso Lorenzo and Lars Lowe Sjösund and Sébastien Cevey and Zach Gleicher and Thi Avrahami and Anudhyan Boral and Hansa Srinivasan and Vittorio Selo and Rhys May and Konstantinos Aisopos and Léonard Hussenot and Livio Baldini Soares and Kate Baumli and Michael B. Chang and Adrià Recasens and Ben Caine and Alexander Pritzel and Filip Pavetic and Fabio Pardo and Anita Gergely and Justin Frye and Vinay Ramasesh and Dan Horgan and Kartikeya Badola and Nora Kassner and Subhrajit Roy and Ethan Dyer and Víctor Campos and Alex Tomala and Yunhao Tang and Dalia El Badawy and Elspeth White and Basil Mustafa and Oran Lang and Abhishek Jindal and Sharad Vikram and Zhitao Gong and Sergi Caelles and Ross Hemsley and Gregory Thornton and Fangxiaoyu Feng and Wojciech Stokowiec and Ce Zheng and Phoebe Thacker and Çağlar Ünlü and Zhishuai Zhang and Mohammad Saleh and James Svensson and Max Bileschi and Piyush Patil and Ankesh Anand and Roman Ring and Katerina Tsihlas and Arpi Vezer and Marco Selvi and Toby Shevlane and Mikel Rodriguez and Tom Kwiatkowski and Samira Daruki and Keran Rong and Allan Dafoe and Nicholas FitzGerald and Keren Gu-Lemberg and Mina Khan and Lisa Anne Hendricks and Marie Pellat and Vladimir Feinberg and James Cobon-Kerr and Tara Sainath and Maribeth Rauh and Sayed Hadi Hashemi and Richard Ives and Yana Hasson and YaGuang Li and Eric Noland and Yuan Cao and Nathan Byrd and Le Hou and Qingze Wang and Thibault Sottiaux and Michela Paganini and Jean-Baptiste Lespiau and Alexandre Moufarek and Samer Hassan and Kaushik Shivakumar and Joost van Amersfoort and Amol Mandhane and Pratik Joshi and Anirudh Goyal and Matthew Tung and Andrew Brock and Hannah Sheahan and Vedant Misra and Cheng Li and Nemanja Rakićević and Mostafa Dehghani and Fangyu Liu and Sid Mittal and Junhyuk Oh and Seb Noury and Eren Sezener and Fantine Huot and Matthew Lamm and Nicola De Cao and Charlie Chen and Gamaleldin Elsayed and Ed Chi and Mahdis Mahdieh and Ian Tenney and Nan Hua and Ivan Petrychenko and Patrick Kane and Dylan Scandinaro and Rishub Jain and Jonathan Uesato and Romina Datta and Adam Sadovsky and Oskar Bunyan and Dominik Rabiej and Shimu Wu and John Zhang and Gautam Vasudevan and Edouard Leurent and Mahmoud Alnahlawi and Ionut Georgescu and Nan Wei and Ivy Zheng and Betty Chan and Pam G Rabinovitch and Piotr Stanczyk and Ye Zhang and David Steiner and Subhajit Naskar and Michael Azzam and Matthew Johnson and Adam Paszke and Chung-Cheng Chiu and Jaume Sanchez Elias and Afroz Mohiuddin and Faizan Muhammad and Jin Miao and Andrew Lee and Nino Vieillard and Sahitya Potluri and Jane Park and Elnaz Davoodi and Jiageng Zhang and Jeff Stanway and Drew Garmon and Abhijit Karmarkar and Zhe Dong and Jong Lee and Aviral Kumar and Luowei Zhou and Jonathan Evens and William Isaac and Zhe Chen and Johnson Jia and Anselm Levskaya and Zhenkai Zhu and Chris Gorgolewski and Peter Grabowski and Yu Mao and Alberto Magni and Kaisheng Yao and Javier Snaider and Norman Casagrande and Paul Suganthan and Evan Palmer and Geoffrey Irving and Edward Loper and Manaal Faruqui and Isha Arkatkar and Nanxin Chen and Izhak Shafran and Michael Fink and Alfonso Castaño and Irene Giannoumis and Wooyeol Kim and Mikołaj Rybiński and Ashwin Sreevatsa and Jennifer Prendki and David Soergel and Adrian Goedeckemeyer and Willi Gierke and Mohsen Jafari and Meenu Gaba and Jeremy Wiesner and Diana Gage Wright and Yawen Wei and Harsha Vashisht and Yana Kulizhskaya and Jay Hoover and Maigo Le and Lu Li and Chimezie Iwuanyanwu and Lu Liu and Kevin Ramirez and Andrey Khorlin and Albert Cui and Tian LIN and Marin Georgiev and Marcus Wu and Ricardo Aguilar and Keith Pallo and Abhishek Chakladar and Alena Repina and Xihui Wu and Tom van der Weide and Priya Ponnapalli and Caroline Kaplan and Jiri Simsa and Shuangfeng Li and Olivier Dousse and Fan Yang and Jeff Piper and Nathan Ie and Minnie Lui and Rama Pasumarthi and Nathan Lintz and Anitha Vijayakumar and Lam Nguyen Thiet and Daniel Andor and Pedro Valenzuela and Cosmin Paduraru and Daiyi Peng and Katherine Lee and Shuyuan Zhang and Somer Greene and Duc Dung Nguyen and Paula Kurylowicz and Sarmishta Velury and Sebastian Krause and Cassidy Hardin and Lucas Dixon and Lili Janzer and Kiam Choo and Ziqiang Feng and Biao Zhang and Achintya Singhal and Tejasi Latkar and Mingyang Zhang and Quoc Le and Elena Allica Abellan and Dayou Du and Dan McKinnon and Natasha Antropova and Tolga Bolukbasi and Orgad Keller and David Reid and Daniel Finchelstein and Maria Abi Raad and Remi Crocker and Peter Hawkins and Robert Dadashi and Colin Gaffney and Sid Lall and Ken Franko and Egor Filonov and Anna Bulanova and Rémi Leblond and Vikas Yadav and Shirley Chung and Harry Askham and Luis C. Cobo and Kelvin Xu and Felix Fischer and Jun Xu and Christina Sorokin and Chris Alberti and Chu-Cheng Lin and Colin Evans and Hao Zhou and Alek Dimitriev and Hannah Forbes and Dylan Banarse and Zora Tung and Jeremiah Liu and Mark Omernick and Colton Bishop and Chintu Kumar and Rachel Sterneck and Ryan Foley and Rohan Jain and Swaroop Mishra and Jiawei Xia and Taylor Bos and Geoffrey Cideron and Ehsan Amid and Francesco Piccinno and Xingyu Wang and Praseem Banzal and Petru Gurita and Hila Noga and Premal Shah and Daniel J. Mankowitz and Alex Polozov and Nate Kushman and Victoria Krakovna and Sasha Brown and MohammadHossein Bateni and Dennis Duan and Vlad Firoiu and Meghana Thotakuri and Tom Natan and Anhad Mohananey and Matthieu Geist and Sidharth Mudgal and Sertan Girgin and Hui Li and Jiayu Ye and Ofir Roval and Reiko Tojo and Michael Kwong and James Lee-Thorp and Christopher Yew and Quan Yuan and Sumit Bagri and Danila Sinopalnikov and Sabela Ramos and John Mellor and Abhishek Sharma and Aliaksei Severyn and Jonathan Lai and Kathy Wu and Heng-Tze Cheng and David Miller and Nicolas Sonnerat and Denis Vnukov and Rory Greig and Jennifer Beattie and Emily Caveness and Libin Bai and Julian Eisenschlos and Alex Korchemniy and Tomy Tsai and Mimi Jasarevic and Weize Kong and Phuong Dao and Zeyu Zheng and Frederick Liu and Fan Yang and Rui Zhu and Mark Geller and Tian Huey Teh and Jason Sanmiya and Evgeny Gladchenko and Nejc Trdin and Andrei Sozanschi and Daniel Toyama and Evan Rosen and Sasan Tavakkol and Linting Xue and Chen Elkind and Oliver Woodman and John Carpenter and George Papamakarios and Rupert Kemp and Sushant Kafle and Tanya Grunina and Rishika Sinha and Alice Talbert and Abhimanyu Goyal and Diane Wu and Denese Owusu-Afriyie and Cosmo Du and Chloe Thornton and Jordi Pont-Tuset and Pradyumna Narayana and Jing Li and Sabaer Fatehi and John Wieting and Omar Ajmeri and Benigno Uria and Tao Zhu and Yeongil Ko and Laura Knight and Amélie Héliou and Ning Niu and Shane Gu and Chenxi Pang and Dustin Tran and Yeqing Li and Nir Levine and Ariel Stolovich and Norbert Kalb and Rebeca Santamaria-Fernandez and Sonam Goenka and Wenny Yustalim and Robin Strudel and Ali Elqursh and Balaji Lakshminarayanan and Charlie Deck and Shyam Upadhyay and Hyo Lee and Mike Dusenberry and Zonglin Li and Xuezhi Wang and Kyle Levin and Raphael Hoffmann and Dan Holtmann-Rice and Olivier Bachem and Summer Yue and Sho Arora and Eric Malmi and Daniil Mirylenka and Qijun Tan and Christy Koh and Soheil Hassas Yeganeh and Siim Põder and Steven Zheng and Francesco Pongetti and Mukarram Tariq and Yanhua Sun and Lucian Ionita and Mojtaba Seyedhosseini and Pouya Tafti and Ragha Kotikalapudi and Zhiyu Liu and Anmol Gulati and Jasmine Liu and Xinyu Ye and Bart Chrzaszcz and Lily Wang and Nikhil Sethi and Tianrun Li and Ben Brown and Shreya Singh and Wei Fan and Aaron Parisi and Joe Stanton and Chenkai Kuang and Vinod Koverkathu and Christopher A. Choquette-Choo and Yunjie Li and TJ Lu and Abe Ittycheriah and Prakash Shroff and Pei Sun and Mani Varadarajan and Sanaz Bahargam and Rob Willoughby and David Gaddy and Ishita Dasgupta and Guillaume Desjardins and Marco Cornero and Brona Robenek and Bhavishya Mittal and Ben Albrecht and Ashish Shenoy and Fedor Moiseev and Henrik Jacobsson and Alireza Ghaffarkhah and Morgane Rivière and Alanna Walton and Clément Crepy and Alicia Parrish and Yuan Liu and Zongwei Zhou and Clement Farabet and Carey Radebaugh and Praveen Srinivasan and Claudia van der Salm and Andreas Fidjeland and Salvatore Scellato and Eri Latorre-Chimoto and Hanna Klimczak-Plucińska and David Bridson and Dario de Cesare and Tom Hudson and Piermaria Mendolicchio and Lexi Walker and Alex Morris and Ivo Penchev and Matthew Mauger and Alexey Guseynov and Alison Reid and Seth Odoom and Lucia Loher and Victor Cotruta and Madhavi Yenugula and Dominik Grewe and Anastasia Petrushkina and Tom Duerig and Antonio Sanchez and Steve Yadlowsky and Amy Shen and Amir Globerson and Adam Kurzrok and Lynette Webb and Sahil Dua and Dong Li and Preethi Lahoti and Surya Bhupatiraju and Dan Hurt and Haroon Qureshi and Ananth Agarwal and Tomer Shani and Matan Eyal and Anuj Khare and Shreyas Rammohan Belle and Lei Wang and Chetan Tekur and Mihir Sanjay Kale and Jinliang Wei and Ruoxin Sang and Brennan Saeta and Tyler Liechty and Yi Sun and Yao Zhao and Stephan Lee and Pandu Nayak and Doug Fritz and Manish Reddy Vuyyuru and John Aslanides and Nidhi Vyas and Martin Wicke and Xiao Ma and Taylan Bilal and Evgenii Eltyshev and Daniel Balle and Nina Martin and Hardie Cate and James Manyika and Keyvan Amiri and Yelin Kim and Xi Xiong and Kai Kang and Florian Luisier and Nilesh Tripuraneni and David Madras and Mandy Guo and Austin Waters and Oliver Wang and Joshua Ainslie and Jason Baldridge and Han Zhang and Garima Pruthi and Jakob Bauer and Feng Yang and Riham Mansour and Jason Gelman and Yang Xu and George Polovets and Ji Liu and Honglong Cai and Warren Chen and XiangHai Sheng and Emily Xue and Sherjil Ozair and Adams Yu and Christof Angermueller and Xiaowei Li and Weiren Wang and Julia Wiesinger and Emmanouil Koukoumidis and Yuan Tian and Anand Iyer and Madhu Gurumurthy and Mark Goldenson and Parashar Shah and MK Blake and Hongkun Yu and Anthony Urbanowicz and Jennimaria Palomaki and Chrisantha Fernando and Kevin Brooks and Ken Durden and Harsh Mehta and Nikola Momchev and Elahe Rahimtoroghi and Maria Georgaki and Amit Raul and Sebastian Ruder and Morgan Redshaw and Jinhyuk Lee and Komal Jalan and Dinghua Li and Ginger Perng and Blake Hechtman and Parker Schuh and Milad Nasr and Mia Chen and Kieran Milan and Vladimir Mikulik and Trevor Strohman and Juliana Franco and Tim Green and Demis Hassabis and Koray Kavukcuoglu and Jeffrey Dean and Oriol Vinyals},
      year={2023},
      eprint={2312.11805},
      archivePrefix={arXiv},
      primaryClass={cs.CL}
}

@misc{aghajanyan2023scaling,
      title={Scaling Laws for Generative Mixed-Modal Language Models}, 
      author={Armen Aghajanyan and Lili Yu and Alexis Conneau and Wei-Ning Hsu and Karen Hambardzumyan and Susan Zhang and Stephen Roller and Naman Goyal and Omer Levy and Luke Zettlemoyer},
      year={2023},
      eprint={2301.03728},
      archivePrefix={arXiv},
      primaryClass={cs.CL}
}

@article{llama3modelcard,
  title={Llama 3 Model Card},
  author={AI@Meta},
  year={2024},
  url = {https://github.com/meta-llama/llama3/blob/main/MODEL_CARD.md}
}

@misc{openai2024gpt4,
      title={GPT-4 Technical Report}, 
      author={OpenAI and : and Josh Achiam and Steven Adler and Sandhini Agarwal and Lama Ahmad and Ilge Akkaya and Florencia Leoni Aleman and Diogo Almeida and Janko Altenschmidt and Sam Altman and Shyamal Anadkat and Red Avila and Igor Babuschkin and Suchir Balaji and Valerie Balcom and Paul Baltescu and Haiming Bao and Mohammad Bavarian and Jeff Belgum and Irwan Bello and Jake Berdine and Gabriel Bernadett-Shapiro and Christopher Berner and Lenny Bogdonoff and Oleg Boiko and Madelaine Boyd and Anna-Luisa Brakman and Greg Brockman and Tim Brooks and Miles Brundage and Kevin Button and Trevor Cai and Rosie Campbell and Andrew Cann and Brittany Carey and Chelsea Carlson and Rory Carmichael and Brooke Chan and Che Chang and Fotis Chantzis and Derek Chen and Sully Chen and Ruby Chen and Jason Chen and Mark Chen and Ben Chess and Chester Cho and Casey Chu and Hyung Won Chung and Dave Cummings and Jeremiah Currier and Yunxing Dai and Cory Decareaux and Thomas Degry and Noah Deutsch and Damien Deville and Arka Dhar and David Dohan and Steve Dowling and Sheila Dunning and Adrien Ecoffet and Atty Eleti and Tyna Eloundou and David Farhi and Liam Fedus and Niko Felix and Simón Posada Fishman and Juston Forte and Isabella Fulford and Leo Gao and Elie Georges and Christian Gibson and Vik Goel and Tarun Gogineni and Gabriel Goh and Rapha Gontijo-Lopes and Jonathan Gordon and Morgan Grafstein and Scott Gray and Ryan Greene and Joshua Gross and Shixiang Shane Gu and Yufei Guo and Chris Hallacy and Jesse Han and Jeff Harris and Yuchen He and Mike Heaton and Johannes Heidecke and Chris Hesse and Alan Hickey and Wade Hickey and Peter Hoeschele and Brandon Houghton and Kenny Hsu and Shengli Hu and Xin Hu and Joost Huizinga and Shantanu Jain and Shawn Jain and Joanne Jang and Angela Jiang and Roger Jiang and Haozhun Jin and Denny Jin and Shino Jomoto and Billie Jonn and Heewoo Jun and Tomer Kaftan and Łukasz Kaiser and Ali Kamali and Ingmar Kanitscheider and Nitish Shirish Keskar and Tabarak Khan and Logan Kilpatrick and Jong Wook Kim and Christina Kim and Yongjik Kim and Jan Hendrik Kirchner and Jamie Kiros and Matt Knight and Daniel Kokotajlo and Łukasz Kondraciuk and Andrew Kondrich and Aris Konstantinidis and Kyle Kosic and Gretchen Krueger and Vishal Kuo and Michael Lampe and Ikai Lan and Teddy Lee and Jan Leike and Jade Leung and Daniel Levy and Chak Ming Li and Rachel Lim and Molly Lin and Stephanie Lin and Mateusz Litwin and Theresa Lopez and Ryan Lowe and Patricia Lue and Anna Makanju and Kim Malfacini and Sam Manning and Todor Markov and Yaniv Markovski and Bianca Martin and Katie Mayer and Andrew Mayne and Bob McGrew and Scott Mayer McKinney and Christine McLeavey and Paul McMillan and Jake McNeil and David Medina and Aalok Mehta and Jacob Menick and Luke Metz and Andrey Mishchenko and Pamela Mishkin and Vinnie Monaco and Evan Morikawa and Daniel Mossing and Tong Mu and Mira Murati and Oleg Murk and David Mély and Ashvin Nair and Reiichiro Nakano and Rajeev Nayak and Arvind Neelakantan and Richard Ngo and Hyeonwoo Noh and Long Ouyang and Cullen O'Keefe and Jakub Pachocki and Alex Paino and Joe Palermo and Ashley Pantuliano and Giambattista Parascandolo and Joel Parish and Emy Parparita and Alex Passos and Mikhail Pavlov and Andrew Peng and Adam Perelman and Filipe de Avila Belbute Peres and Michael Petrov and Henrique Ponde de Oliveira Pinto and Michael and Pokorny and Michelle Pokrass and Vitchyr H. Pong and Tolly Powell and Alethea Power and Boris Power and Elizabeth Proehl and Raul Puri and Alec Radford and Jack Rae and Aditya Ramesh and Cameron Raymond and Francis Real and Kendra Rimbach and Carl Ross and Bob Rotsted and Henri Roussez and Nick Ryder and Mario Saltarelli and Ted Sanders and Shibani Santurkar and Girish Sastry and Heather Schmidt and David Schnurr and John Schulman and Daniel Selsam and Kyla Sheppard and Toki Sherbakov and Jessica Shieh and Sarah Shoker and Pranav Shyam and Szymon Sidor and Eric Sigler and Maddie Simens and Jordan Sitkin and Katarina Slama and Ian Sohl and Benjamin Sokolowsky and Yang Song and Natalie Staudacher and Felipe Petroski Such and Natalie Summers and Ilya Sutskever and Jie Tang and Nikolas Tezak and Madeleine B. Thompson and Phil Tillet and Amin Tootoonchian and Elizabeth Tseng and Preston Tuggle and Nick Turley and Jerry Tworek and Juan Felipe Cerón Uribe and Andrea Vallone and Arun Vijayvergiya and Chelsea Voss and Carroll Wainwright and Justin Jay Wang and Alvin Wang and Ben Wang and Jonathan Ward and Jason Wei and CJ Weinmann and Akila Welihinda and Peter Welinder and Jiayi Weng and Lilian Weng and Matt Wiethoff and Dave Willner and Clemens Winter and Samuel Wolrich and Hannah Wong and Lauren Workman and Sherwin Wu and Jeff Wu and Michael Wu and Kai Xiao and Tao Xu and Sarah Yoo and Kevin Yu and Qiming Yuan and Wojciech Zaremba and Rowan Zellers and Chong Zhang and Marvin Zhang and Shengjia Zhao and Tianhao Zheng and Juntang Zhuang and William Zhuk and Barret Zoph},
      year={2024},
      eprint={2303.08774},
      archivePrefix={arXiv},
      primaryClass={cs.CL}
}

@misc{sharma2020neural,
      title={A Neural Scaling Law from the Dimension of the Data Manifold}, 
      author={Utkarsh Sharma and Jared Kaplan},
      year={2020},
      eprint={2004.10802},
      archivePrefix={arXiv},
      primaryClass={cs.LG}
}

@misc{bahri2021explaining,
      title={Explaining Neural Scaling Laws}, 
      author={Yasaman Bahri and Ethan Dyer and Jared Kaplan and Jaehoon Lee and Utkarsh Sharma},
      year={2021},
      eprint={2102.06701},
      archivePrefix={arXiv},
      primaryClass={cs.LG}
}

@misc{brown2020language,
      title={Language Models are Few-Shot Learners}, 
      author={Tom B. Brown and Benjamin Mann and Nick Ryder and Melanie Subbiah and Jared Kaplan and Prafulla Dhariwal and Arvind Neelakantan and Pranav Shyam and Girish Sastry and Amanda Askell and Sandhini Agarwal and Ariel Herbert-Voss and Gretchen Krueger and Tom Henighan and Rewon Child and Aditya Ramesh and Daniel M. Ziegler and Jeffrey Wu and Clemens Winter and Christopher Hesse and Mark Chen and Eric Sigler and Mateusz Litwin and Scott Gray and Benjamin Chess and Jack Clark and Christopher Berner and Sam McCandlish and Alec Radford and Ilya Sutskever and Dario Amodei},
      year={2020},
      eprint={2005.14165},
      archivePrefix={arXiv},
      primaryClass={cs.CL}
}

@misc{villalobos2022run,
      title={Will we run out of data? An analysis of the limits of scaling datasets in Machine Learning}, 
      author={Pablo Villalobos and Jaime Sevilla and Lennart Heim and Tamay Besiroglu and Marius Hobbhahn and Anson Ho},
      year={2022},
      eprint={2211.04325},
      archivePrefix={arXiv},
      primaryClass={cs.LG}
}

\appendix
\section{More details about the scaling law for transfer}
\label{app:derive_scaling_law}
\begin{figure}[h]
    \centering
    \includegraphics[width=0.8\textwidth]{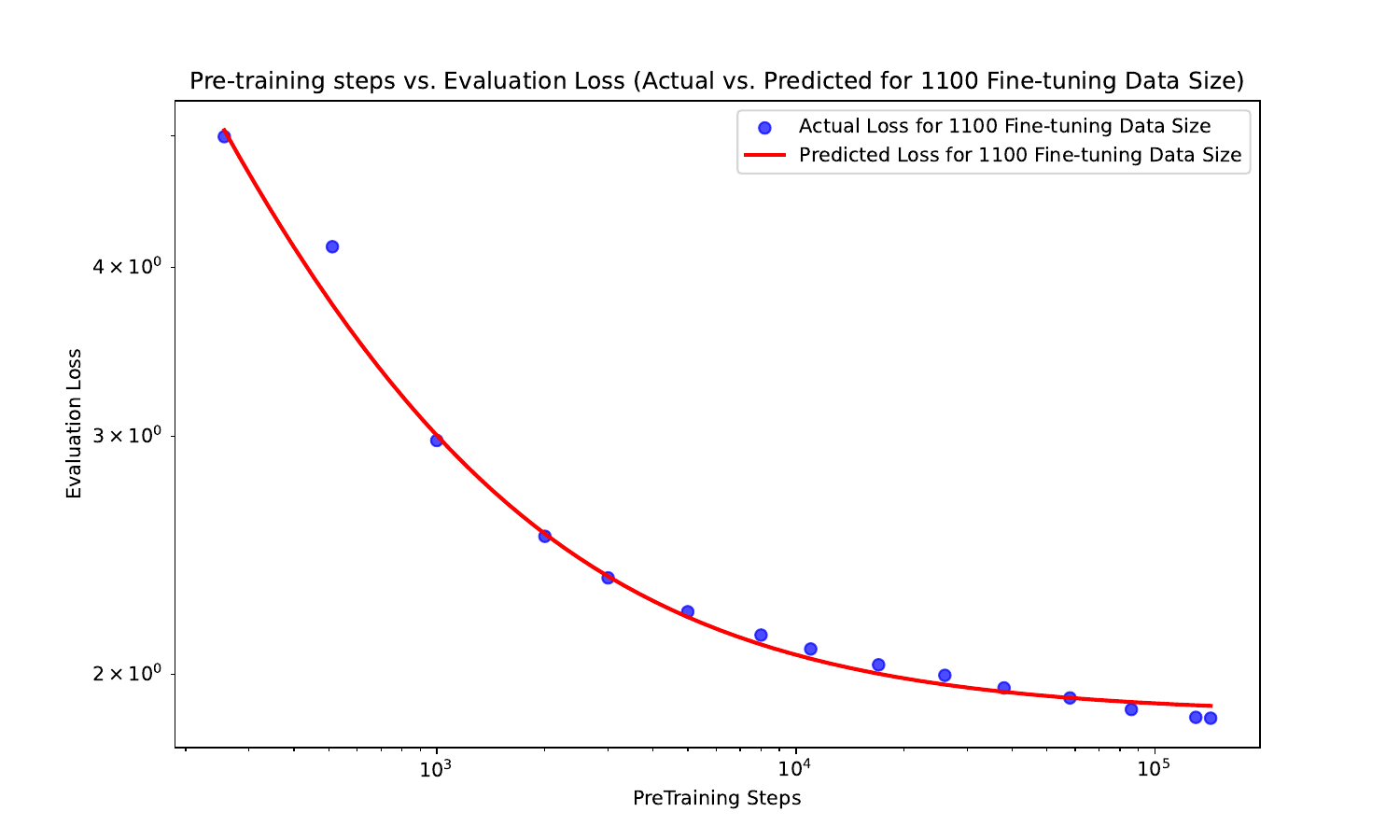}
    \caption{This plot illustrates a cross-section of the fitted scaling law to the data for the fictional encyclopedia dataset, illustrating both clear transfer learning, and that the power law form provides a good fit in pre-training data steps. This empirical observation confirms our intuitions that the scaling law for transfer should reduce to a power law under various conditions. These conditions are detailed in \autoref{app:derive_scaling_law}}.
    \label{fig:Plots/power_law_loss}
\end{figure}

To speculate the form of the scaling law for transfer, we borrow from \cite{mikami2021scaling}, who proposed plausible criteria for such a law, derived from their empirical analysis. It is important to note some key differences between their study and ours: they focused on \(L_2\) loss in images, whereas our study deals with cross-entropy loss on mainly natural language data. Additionally, their analysis used the number of fine-tuning data points seen as the data variable, in contrast to our use of fine-tuning data size, as we train our models to convergence on the fine-tuning data. Keeping these distinctions in mind, we now revisit the conditions from \cite{mikami2021scaling}:
\begin{enumerate}
    \item \textbf{(Irreducible error)} 
    \begin{align*}
        \lim_{{f \to \infty}} L(p, f) &= E
    \end{align*}
    \item \textbf{(Power law with maximum pre-training)} 
    \begin{align*}
        \lim_{{p \to \infty}} L(p, f) &= G \cdot f^{-\beta} + E
    \end{align*}
    \item \textbf{(Power law with no pre-training)} 
    \begin{align*}
        L(0, f) &= C \cdot f^{-\beta} + E
    \end{align*}
\end{enumerate}
In addition to the three conditions from \cite{mikami2021scaling}, we introduce a fourth condition, which enhances our interpretation of the transfer gap:
\begin{enumerate}    \setcounter{enumi}{3}
    \item \textbf{(Power law plus transfer gap with no fine-tuning)}
    \begin{align*}
        L(p, 0) &= A \cdot p^{-\alpha} + G + E
    \end{align*}
\end{enumerate}
Conditions 2, 3, and 4 are natural assumptions given that the power law form is ubiquitous in machine learning scaling laws (\cite{henighan2020scaling}, \cite{sharma2020neural}, \cite{villalobos2022run}). The empirical strength of the power law form is evidenced by the clear power law-like shape observed in the data. This shape is illustrated by \autoref{fig:Plots/power_law_loss}, which reveals how loss decreases in pre-training steps, for a fixed number of fine-tuning data points.
Given the first three conditions, \cite{mikami2021scaling} speculate the following form, which is not the only possible form that satisfies these four conditions:
\begin{equation}
     L(p, f) = (G + A \cdot p^{-\alpha}) \cdot f^{-\beta} + E
     \label{eq:scaling_law_1}
\end{equation}
\begin{equation}
    = A \cdot p^{-\alpha} \cdot f^{-\beta} + G \cdot f^{-\beta} + E
\end{equation}
Here, $p$ refers to the pre-training data steps, and $f$ refers to the fine-tuning data size, with $A, G, E, \alpha, \beta$ representing the constants in the scaling law, which are determined empirically by fitting the model to data. As previously noted in \autoref{app:scaling_laws}, this scaling law is technically an approximation of the true scaling law, in which the terms $p$ and $f$ are modified to $p-1$ and $f-1$ respectively. This adjustment is necessary to ensure that conditions 3 and 4 of the scaling law are satisfied.
\begin{equation}
L(p, f) = (A \cdot (p+1)^{-\alpha} + G) \cdot (f+1)^{-\beta} + E
\end{equation}
\cite{mikami2021scaling} support their speculation with a theoretical analysis of the evolution of loss during training within the Neural Tangent Kernel (NTK) framework, building on the work of \cite{arora2019finegrained} and \cite{nitanda2020gradient}. However, their theoretical analysis is not directly applicable to the present study, as our fine-tuning data term, $f$, represents the number of fine-tuning data points, rather than the number of steps during training. Consequently, the NTK framework has limited applicability to our analysis.

Instead of relying on theory to justify the scaling law form represented by \autoref{eq:scaling_law_1}, we utilize empirical analysis. Specifically, we explored many alternative plausible forms for the scaling law that satisfy at least some of these four conditions, and rigorously tested their ability to predict our experimental data via extensive cross-validation, as described in \autoref{app:comparison_scaling_laws}.

Despite the various forms we considered, none substantially improved upon the simple form proposed by \cite{mikami2021scaling}, lending credence to the idea that this form is relatively robust. Given the form’s simplicity and good performance, it is the primary form we employed to demonstrate our results.

\subsection{Comparing to \cite{hernandez2021scaling}}

It is useful to contrast our scaling law form with that in \cite{hernandez2021scaling}. In their study, they define \(D_T\), for effective data transferred, as the amount of additional fine-tuning data that a model of the same size trained only on fine-tuning data would have needed to achieve the same loss on fine-tuning data as a model trained purely on pre-training data. This concept is given by the following equation:
\begin{equation}
    D_T = \text{effective data transferred} = k(D_F)^\alpha (N)^\beta
\end{equation}
where $N$ is the number of non-embedding parameters, and $D_F$ is the size of the fine-tuning data. To compare this scaling law with the one from \cite{mikami2021scaling}, we can derive a notion of effective data transferred using \eqref{eq:scaling_law_1} by solving the following equation for $f_2$:
\begin{equation}
    L(p, f_1) = L(0, f_2),
\end{equation}
which indicates that the value for $f_2$ is such that we achieve the loss $L(p, f_1)$ without any pre-training data. Recall that $p = \text{pre-training tokens seen} + 1$.
\begin{equation}
    (G + A \cdot p^{-\alpha}) \cdot f_1^{-\beta} + E = (G + A) \cdot f_2^{-\beta} + E
\end{equation}
\begin{equation}
    f_2 = \left( \frac{f_1^{\beta} \cdot (A + G \cdot p^{-\alpha})}{A + G} \right)^{\frac{1}{\beta}}
\end{equation}
Given the complexity of this functional form, we prefer to refer to \autoref{eq:scaling_law_1} directly as a function of pre-training data processed and fine-tuning datapoints, rather than in terms of effective data transferred as considered by \cite{hernandez2021scaling}.
\newpage
\section{A presentation and analysis of the empirical data}
\label{data_presentation_results}
\begin{figure}[p]
    \centering
    \begin{subfigure}{0.5\textwidth}
        \centering
        \includegraphics[width=\linewidth]{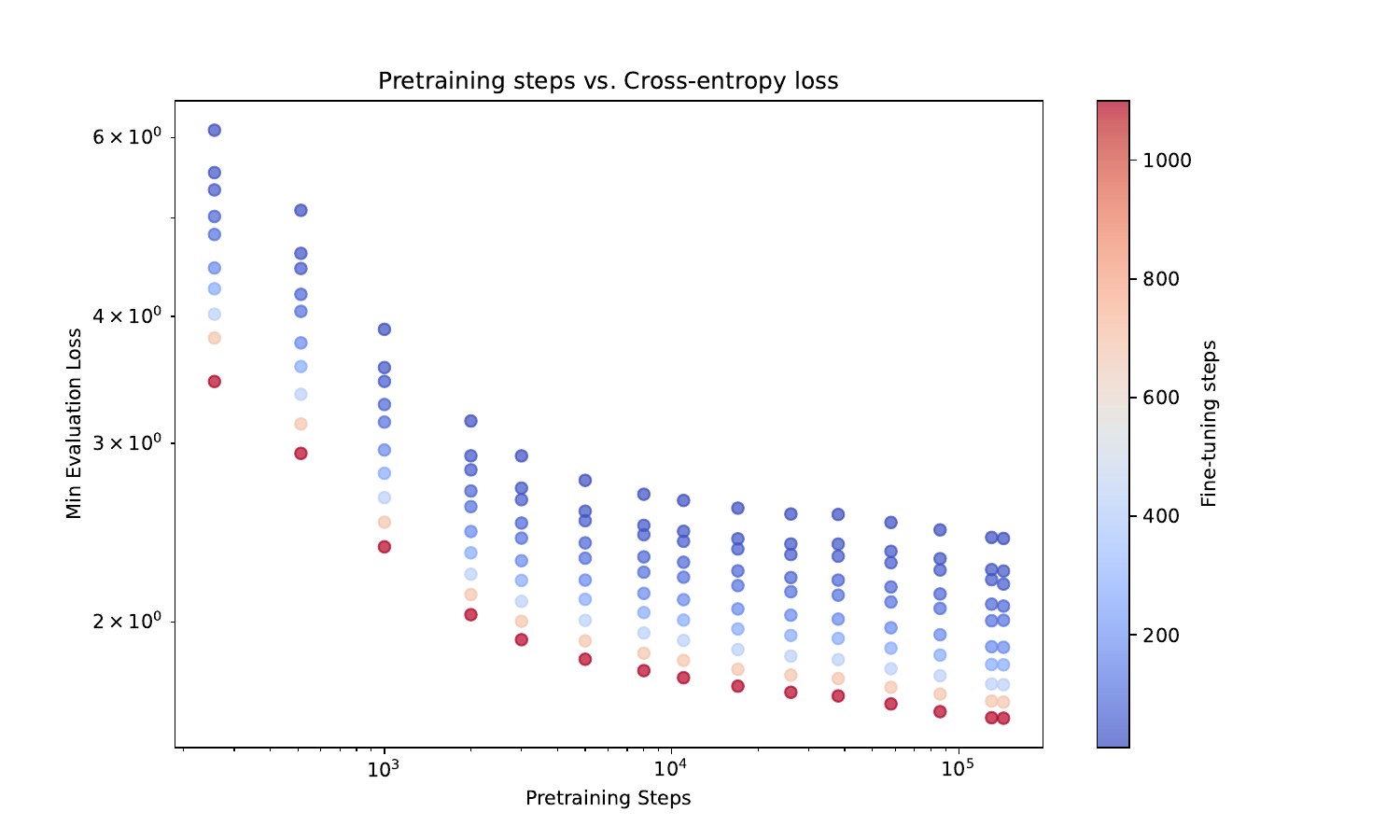}
        \caption{Fictional encyclopedia raw data.}
        \label{fig:Plots/basic_data_1}
    \end{subfigure}%
    \begin{subfigure}{0.5\textwidth}
        \centering
        \includegraphics[width=\linewidth]{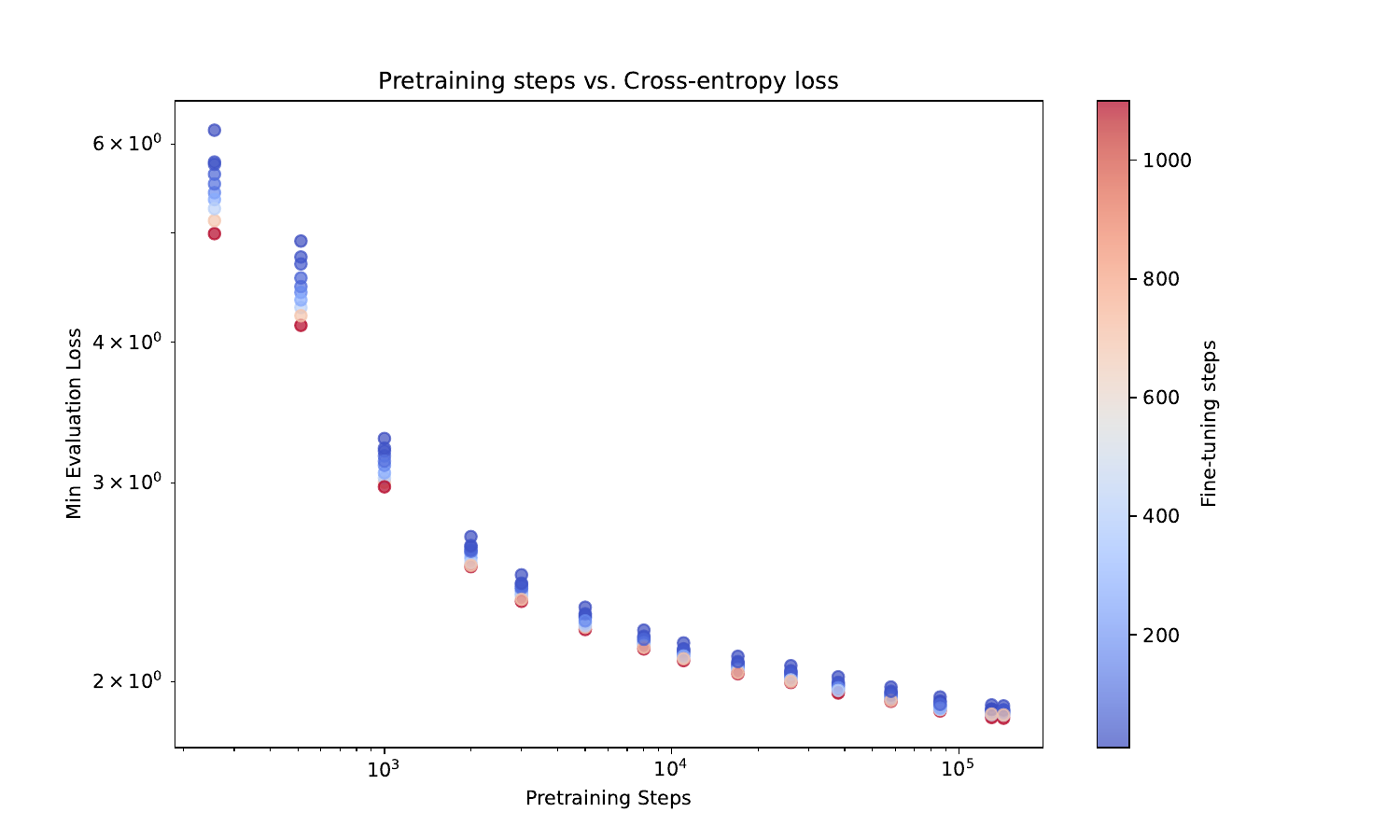}
        \caption{Math arXiv raw data.}
        \label{fig:Plots/basic_data_2}
    \end{subfigure}
    \begin{subfigure}{0.5\textwidth}
        \centering
        \includegraphics[width=\linewidth]{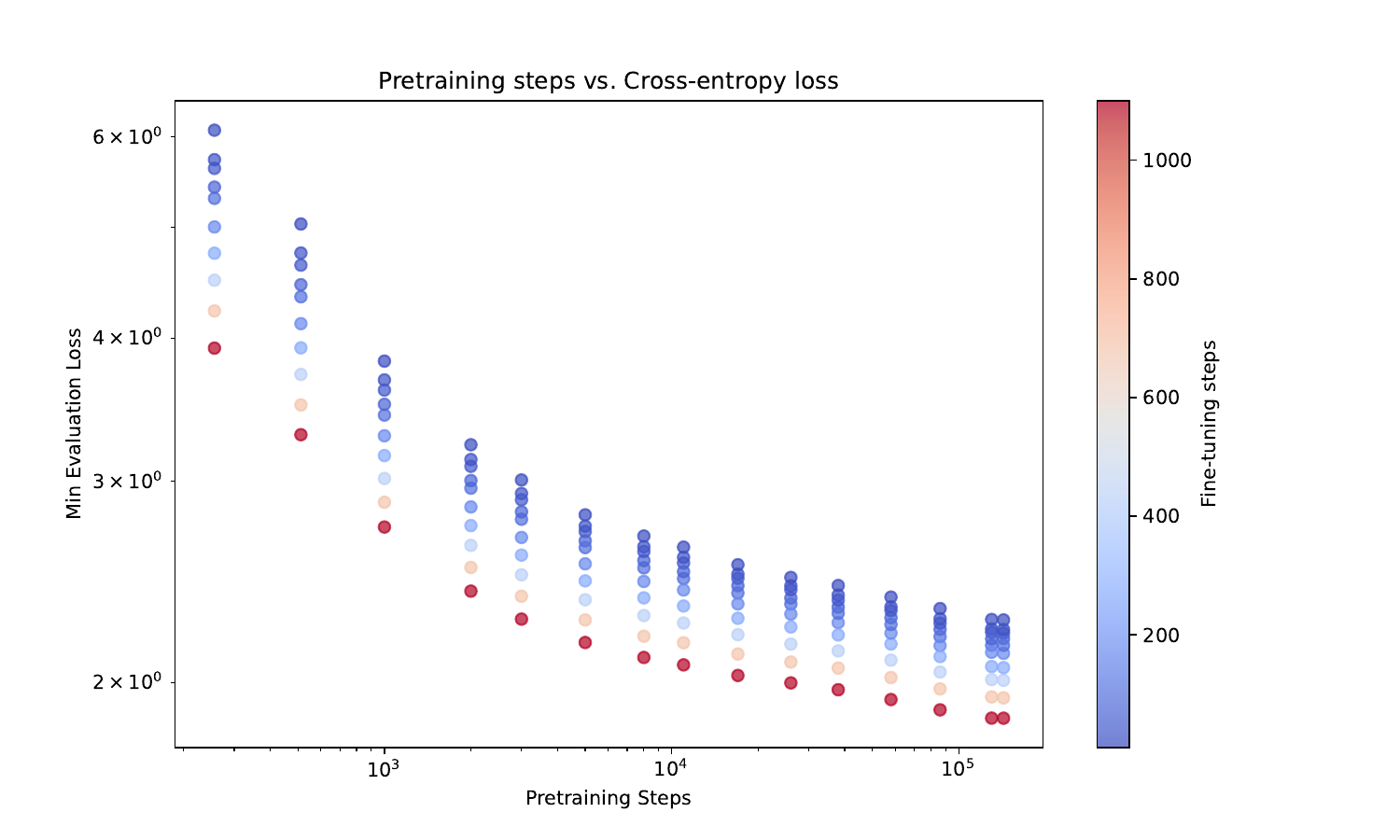}
        \caption{Statistics textbook raw data.}
        \label{fig:Plots/basic_data_3}
    \end{subfigure}%
    \begin{subfigure}{0.5\textwidth}
        \centering
        \includegraphics[width=\linewidth]{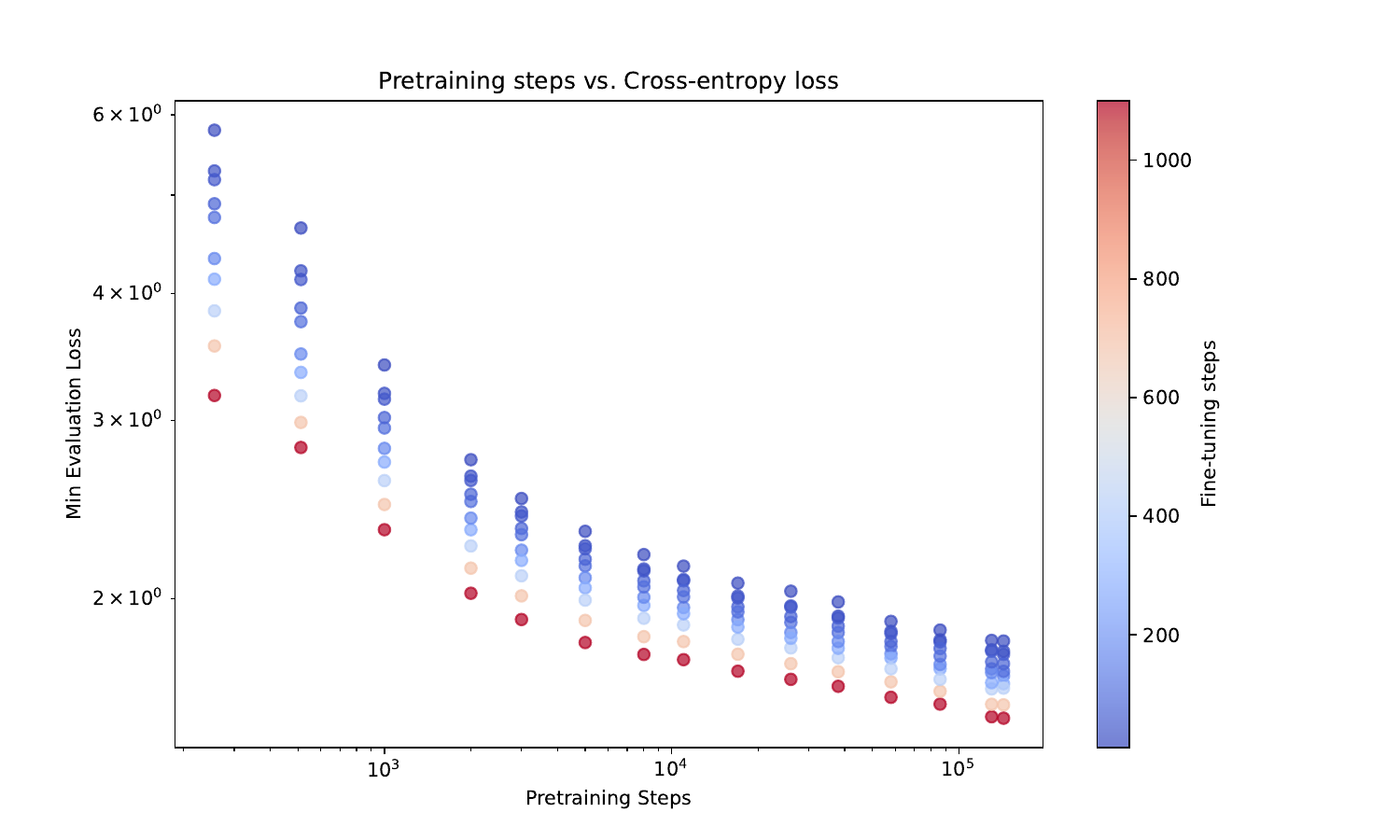}
        \caption{Enron emails raw data.}
        \label{fig:Plots/basic_data_4}
    \end{subfigure}
    \begin{subfigure}{0.5\textwidth}
        \centering
        \includegraphics[width=\linewidth]{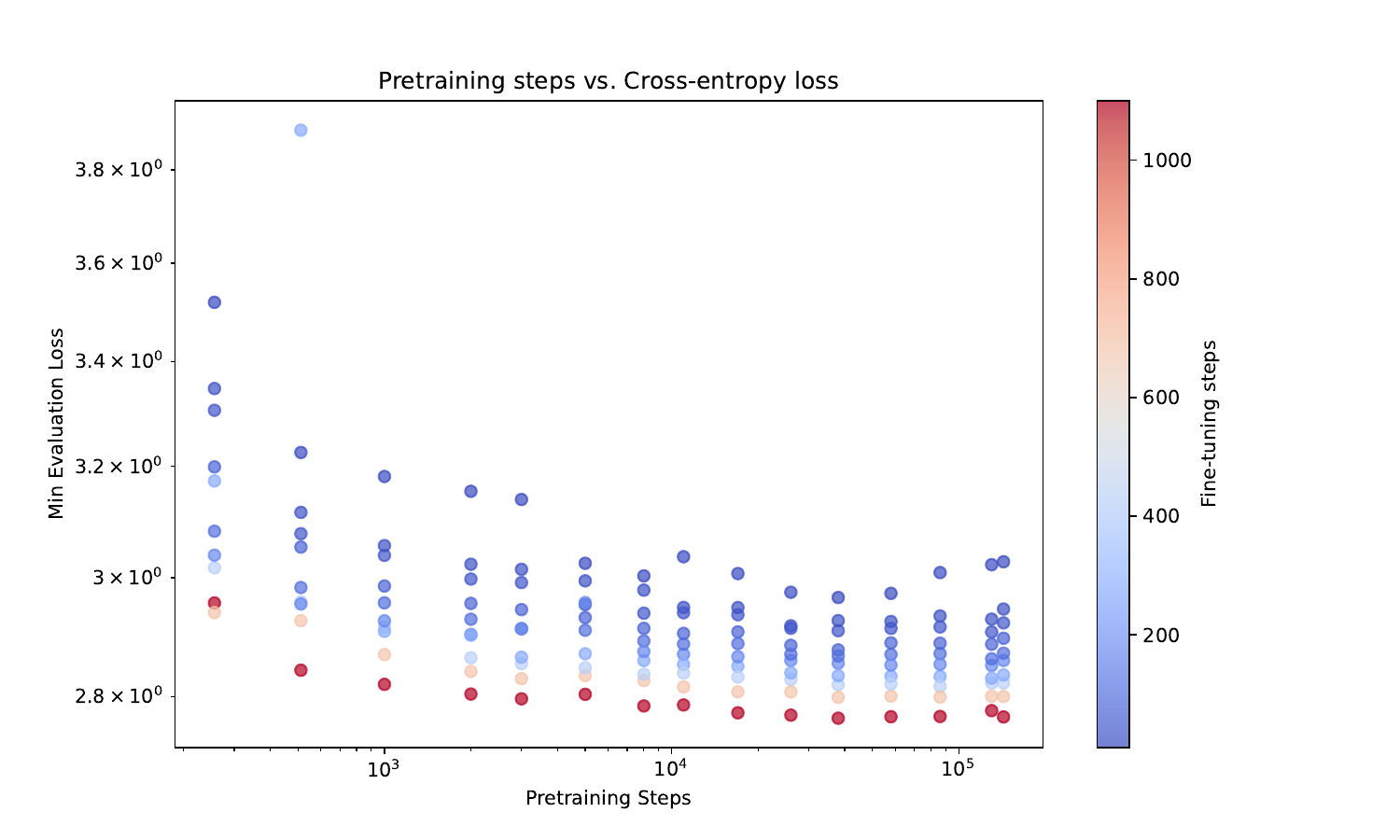}
        \caption{House cat genome raw data.}
        \label{fig:Plots/basic_data_5}
    \end{subfigure}
    \caption{Cross entropy loss as a function of pre-training data steps and fine-tuning data points across all datasets in the study. The plots show clear transfer learning, with decreasing loss with increasing pre-training values. The math arXiv dataset shows slowly decreasing loww from fine-tuning relative to pre-training, reflected in the scaling law by a relatively small fine-tuning exponent. The housecat genome data displays the most erratic and unclear pattern, likely reflecting instability caused from the high dissimilarity of this dataset from the pre-training data.}
    \label{fig:Plots/basic_data_plots}
\end{figure}

A selection of in-depth results from the experiments is presented in this section. A basic presentation of the data is shown in \autoref{fig:Plots/basic_data_plots}. Model performance across varying pre-training steps and fine-tuning data sizes demonstrates clear transfer learning across all datasets, as evidenced by the smoothly decreasing loss with increasing pre-training data for all datasets, with the notable exception of the house cat genome dataset. This dataset exhibits a more erratic pattern compared to the others, likely reflecting its significant dissimilarity to the pile.
\begin{figure}[!htbp]
    \centering
    \includegraphics[width=0.8\textwidth]{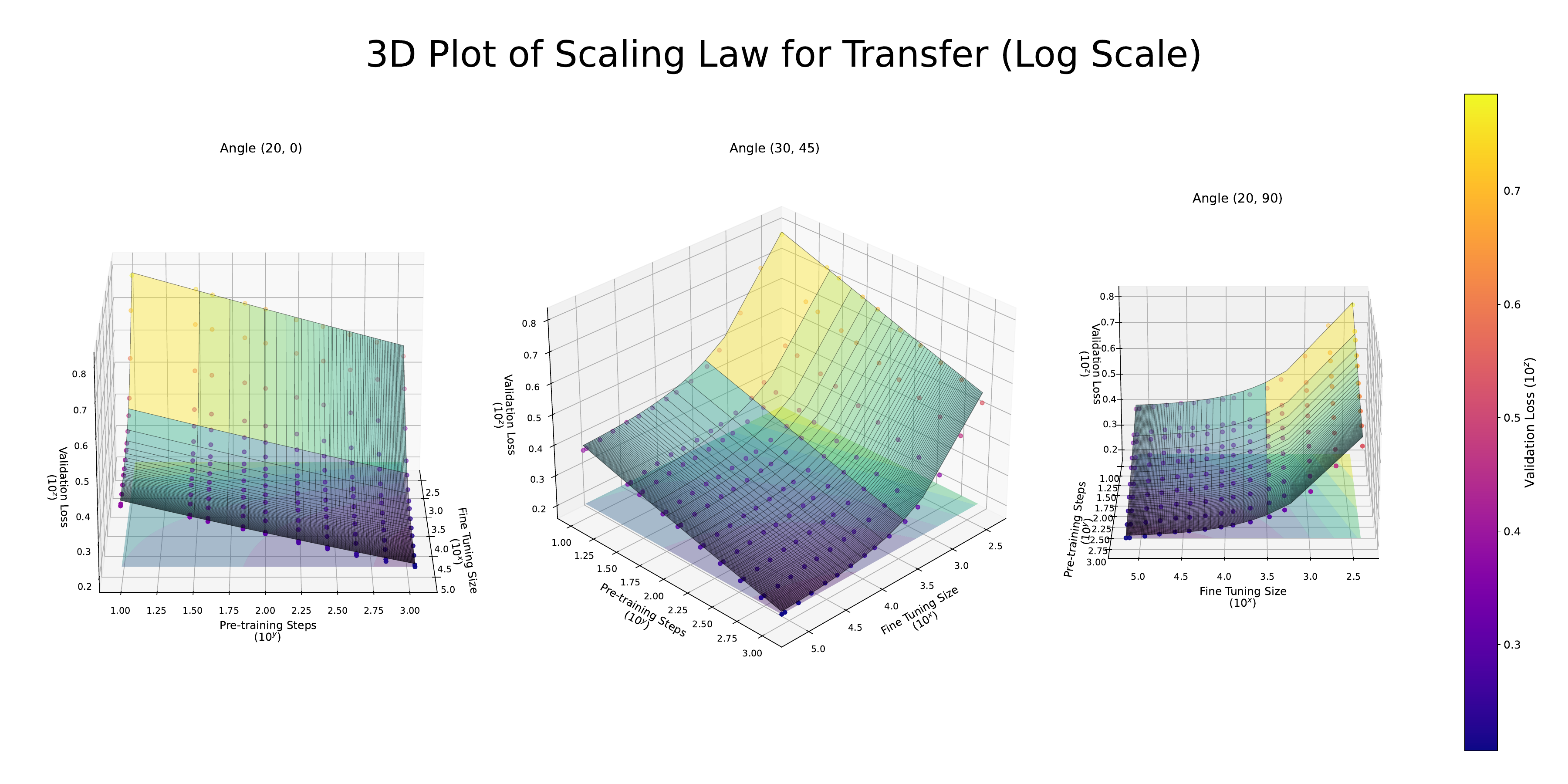}
    \caption{Validation loss compared against the predicted values of the fitted scaling law in three dimensions, in logarithmic space, for the synthetic dataset. From visual inspection, the scaling law appears to be a close fit for the data, fitting the shape of the data points, and showing no obvious signs of overfitting.}
    \label{fig:Plots/3d_log_first}
\end{figure}

\autoref{fig:Plots/3d_log_first} illustrates how well the scaling law (\autoref{eq:scaling_law_1}) fits the experimental data from the fictional encyclopedia dataset, in a 3D plot. Through close visual inspection, it is clear that the underlying structure of the data is well-modeled by the fitted scaling law. In particular, the power law-like relationship in both pre-training steps and fine-tuning data points is well-captured by the scaling law. Given the simple nature of the scaling law, it is unlikely to be overfitting the data significantly. The goodness of fit of this particular scaling law is further validated by the results from cross-validation, outlined in \autoref{app:comparison_scaling_laws}.

\begin{figure}[p]
    \centering
    \begin{subfigure}{0.5\textwidth}
        \centering
        \includegraphics[width=\linewidth]{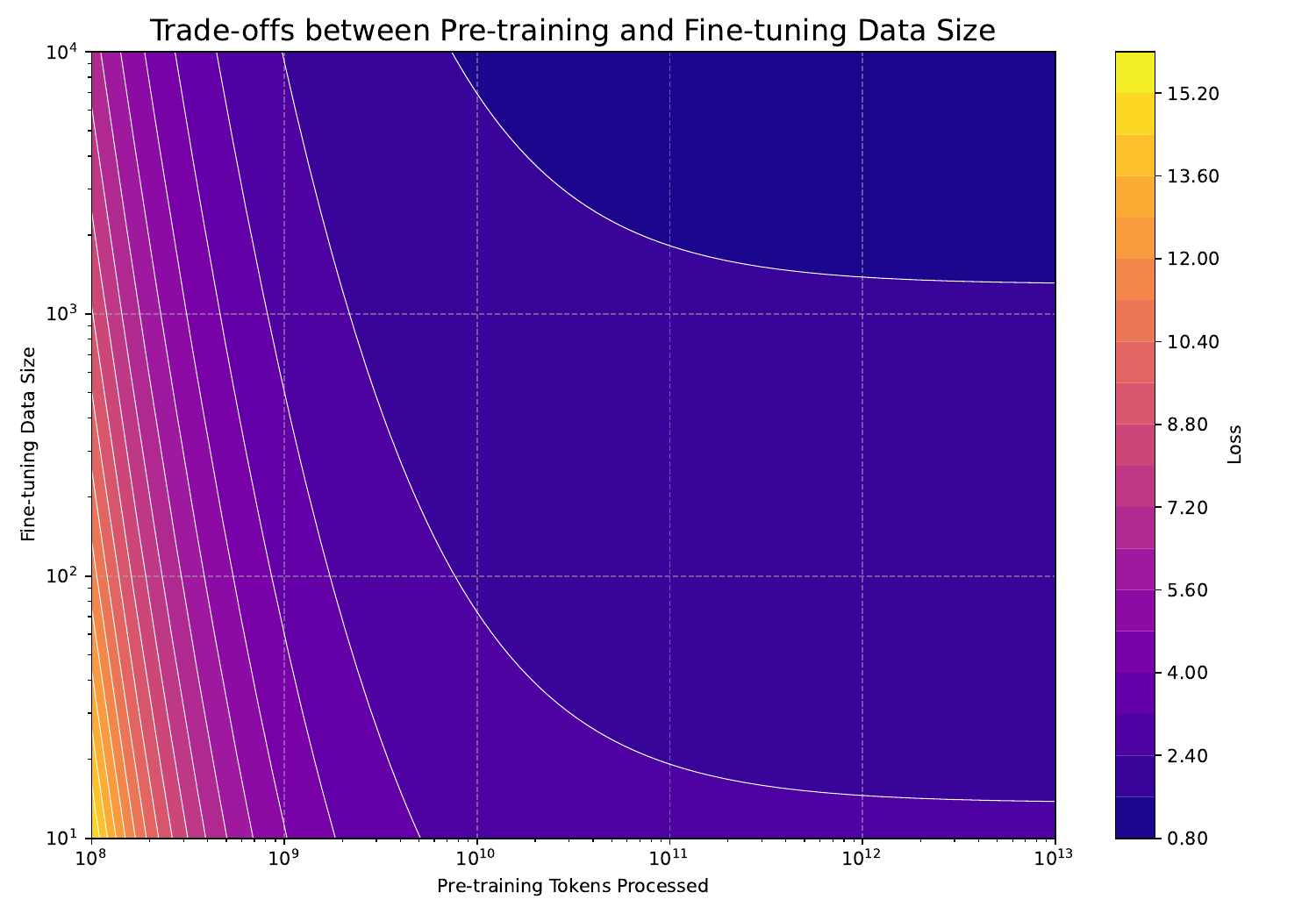}
        \caption{Fictional encyclopedia iso-loss.}
        \label{fig:Plots/iso_loss_1}
    \end{subfigure}%
    \begin{subfigure}{0.5\textwidth}
        \centering
        \includegraphics[width=\linewidth]{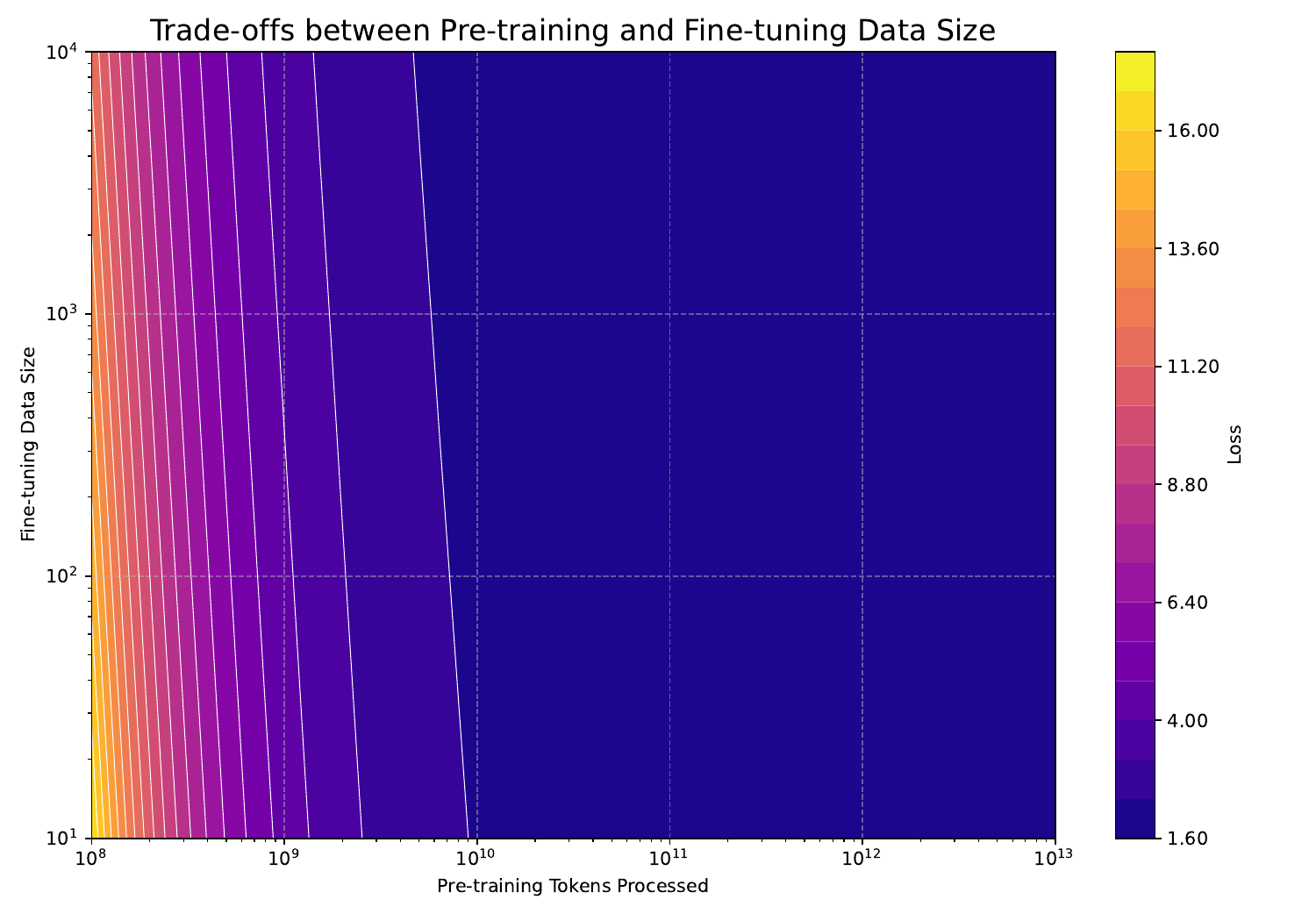}
        \caption{Math arXiv iso-loss.}
        \label{fig:Plots/iso_loss_2}
    \end{subfigure}
    \begin{subfigure}{0.5\textwidth}
        \centering
        \includegraphics[width=\linewidth]{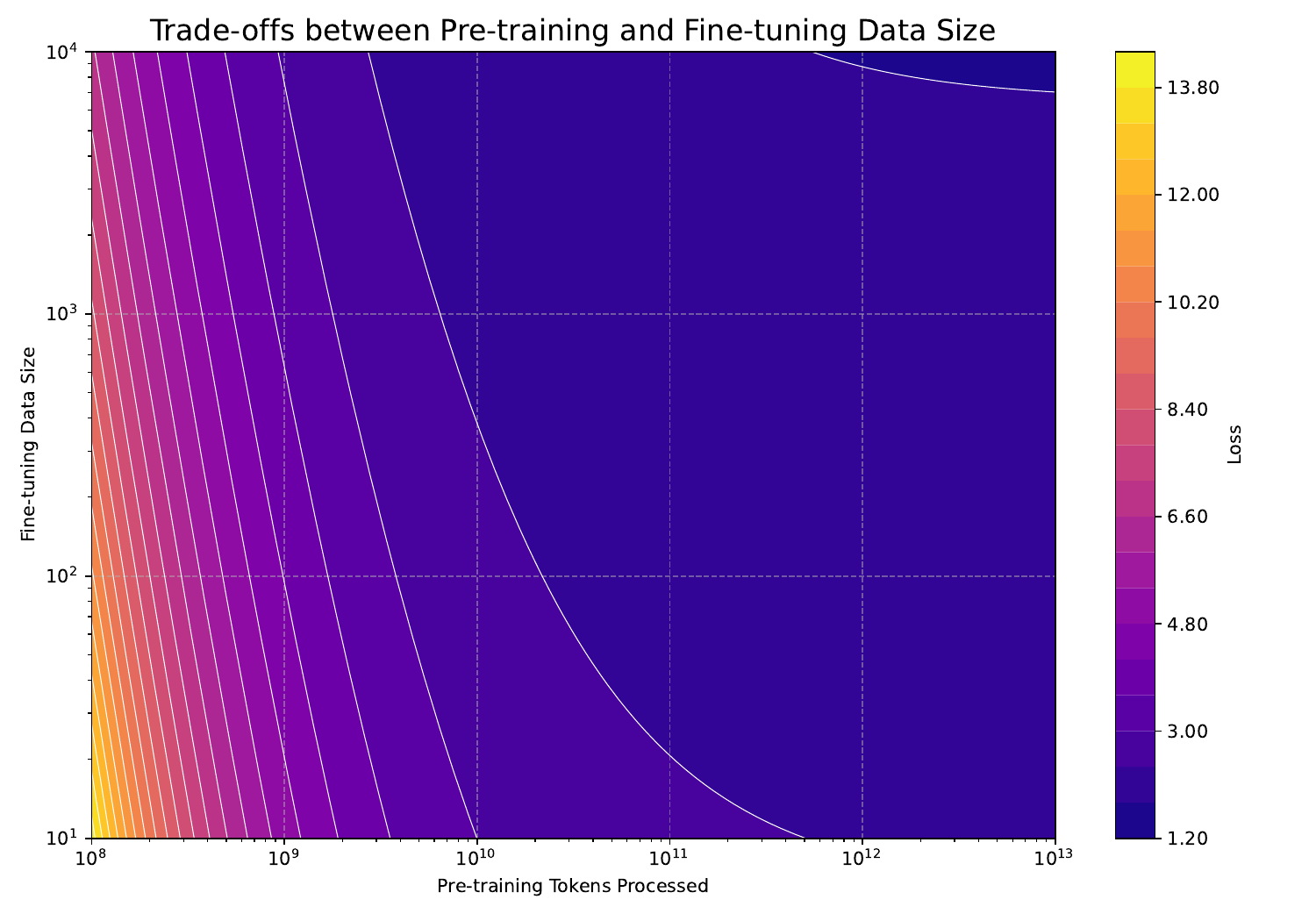}
        \caption{Statistics textbook iso-loss.}
        \label{fig:Plots/iso_loss_3}
    \end{subfigure}%
    \begin{subfigure}{0.5\textwidth}
        \centering
        \includegraphics[width=\linewidth]{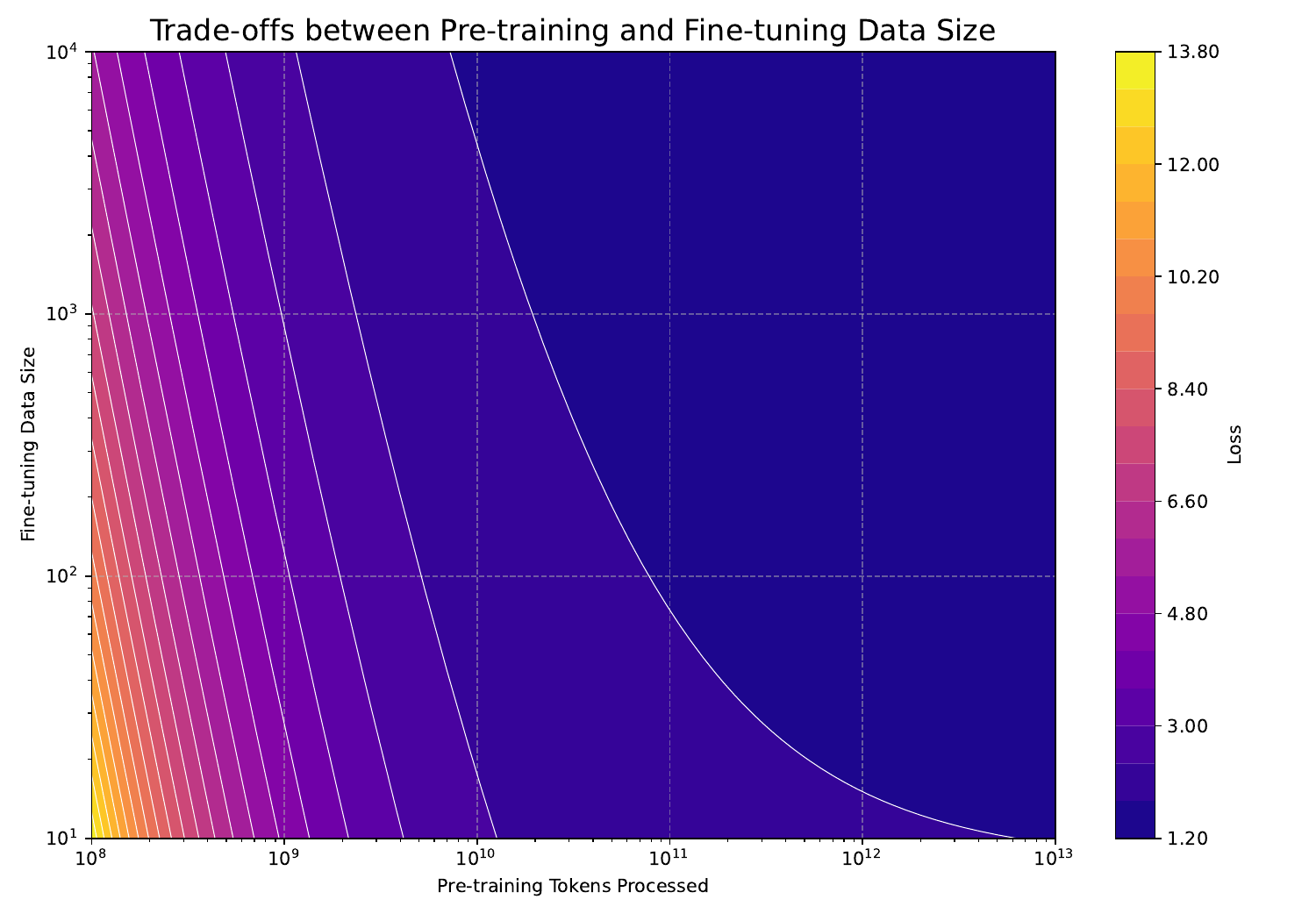}
        \caption{Enron emails iso-loss.}
        \label{fig:Plots/iso_loss_4}
    \end{subfigure}
    \begin{subfigure}{0.5\textwidth}
        \centering
        \includegraphics[width=\linewidth]{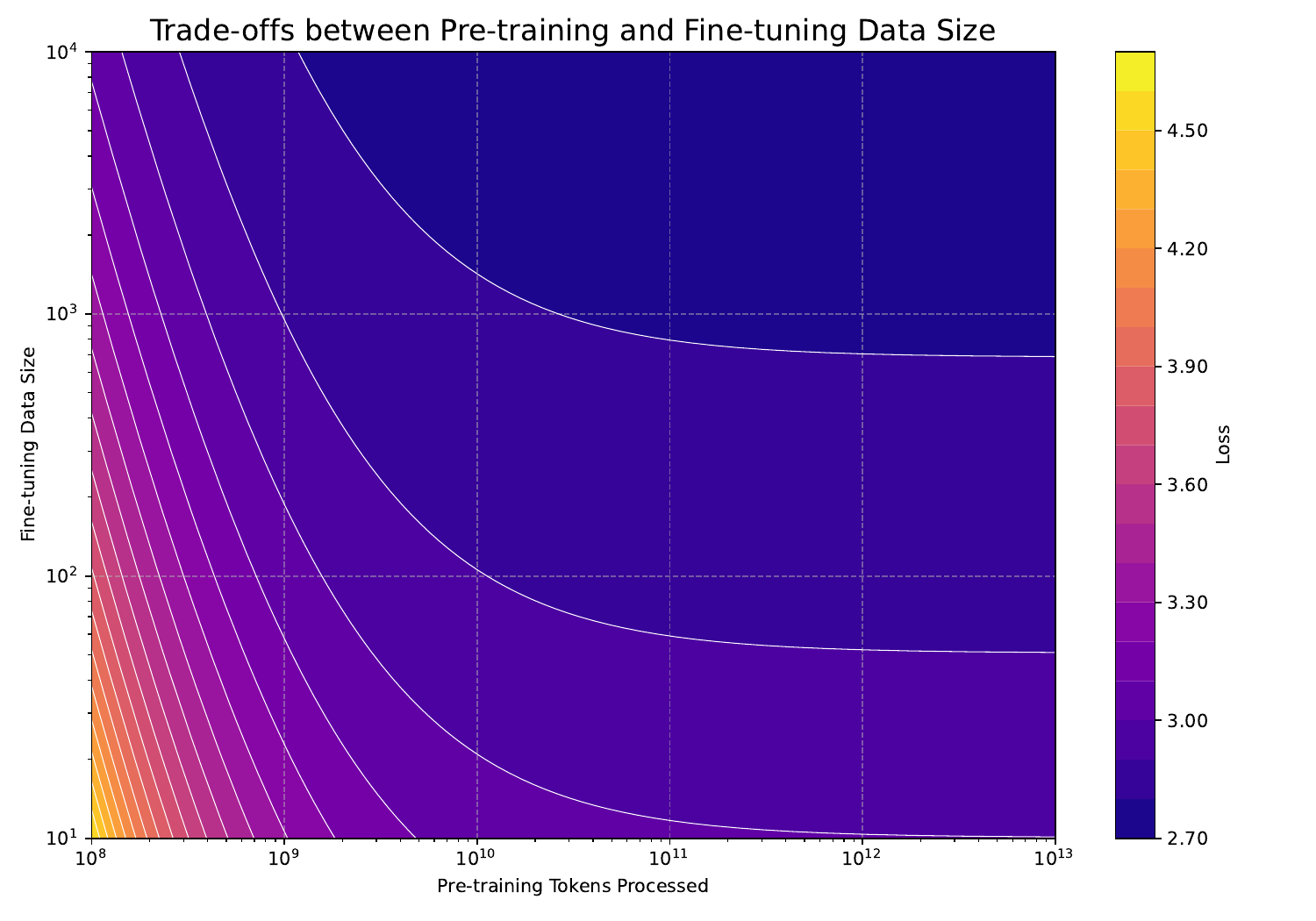}
        \caption{House cat genome iso-loss.}
        \label{fig:Plots/iso_loss_5}
    \end{subfigure}
    \caption{These iso-loss curves illustrate the various ways of achieving identical cross entropy loss using different combinations of training inputs, for each dataset. It is clear that the math arXiv dataset benefits greatly from pre-training relative to fine-tuning, as the iso-loss curves are very steep vertically. By contrast, the fictional encyclopedia dataset benefits greatly from pre-training only in a regime of low pre-training. In a regime of greater pre-training, fine-tuning is a more effective strategy of reducing loss, owing to the nearly horizontal iso-loss curves.}
    \label{fig:Plots/iso_loss_plots}
\end{figure}

Finally, \autoref{fig:Plots/iso_loss_plots} illustrates the trade-offs between expanding pre-training and collecting more fine-tuning data across all of the datasets. Comparing these different plots reveals how optimal strategies for allocating data can change depending on the scaling law for transfer. For example, for the math arXiv dataset, pre-training remains a relatively effective strategy for reducing loss for even high levels of pre-training, as indicated by the nearly-vertical iso-lines. By contrast, for the fictional encyclopedia dataset, pre-training is generally highly beneficial in a regime of low pre-training, but becomes relatively less helpful in a regime of high pre-training, as the isolines become closer to being horizontal. This demonstrates how scaling laws for transfer can be used to inform strategies for allocating data and compute when training models.
\newpage
\section{More information about the experimental methodology}
\begin{table}[ht!]
\centering
\begin{tabular}{ccc}
\toprule
Tokens seen during pre-training & Fine-tuning data size, in tokens \\
\midrule
$5.37 \times 10^{8}$ & $10$ \\
$1.07 \times 10^{9}$ & $30$ \\
$2.10 \times 10^{9}$ & $40$ \\
$4.19 \times 10^{9}$ & $70$ \\
$6.29 \times 10^{9}$ & $100$ \\
$1.05 \times 10^{10}$ & $170$ \\
$1.68 \times 10^{10}$ & $270$ \\
$2.31 \times 10^{10}$ & $430$ \\
$3.57 \times 10^{10}$ & $690$ \\
$5.45 \times 10^{10}$ & $1100$ \\
$7.97 \times 10^{10}$ &  \\
$1.22 \times 10^{11}$ &  \\
$1.80 \times 10^{11}$ &  \\
$2.73 \times 10^{11}$ &  \\
$2.99 \times 10^{11}$ &  \\
\bottomrule
\end{tabular}
\caption{This table details the set of training run experiments for each fine-tuning dataset. Every combination of these training settings represents a single experiment, with a total of 150 experiments per fine-tuning dataset, and a total of 750 experiments across all 5 datasets. The list of fine-tuning datasets can be found in \autoref{tab:datasets}}
\label{tab:experiments}
\end{table}
For each fine-tuning dataset, a set of training runs were performed for the 2.8 billion parameter Pythia model at various pre-training checkpoints. When the model was fit to convergence, the final minimum evaluation loss value was recorded, along with the pre-training tokens seen, fine-tuning data size, and number of epochs until convergence. This comprehensive suite of training runs form the data points used to fit the scaling law for transfer, with each training run representing a single data point. The full set of training runs specifications are provided by \autoref{tab:experiments}.
\subsection{Methodology for fitting the scaling law}
\label{app:fitting_methodology}
To find the parameters presented in \autoref{tab:fitted_parameters}, we employed the BFGS optimization method to solve this minimization problem:
\begin{equation}
\begin{aligned}
    & \underset{a, g, e, \alpha, \beta}{\text{minimize}} \\
    & \sum_{\text{run}_i} \text{Huber}_\delta \bigg ( \log\left(\exp\left(-\beta \log(f_i) + \log\left(\exp(a)p_i^{-\alpha} + \exp(g)\right)\right) + \exp(e)\right) - \log(L_i)\bigg )
\end{aligned}
\end{equation}
where $A=\exp(a), G=\exp(g), E=\exp(e)$. The optimization iteratively refined its estimates, starting from the grid of initial guesses, \(a \in \{0.0, 2.0, \ldots, 8.0\}\), \(b \in \{0.0, 0.33, \ldots, 1.0\}\), \(c \in \{-5.0, -3.33, \ldots, 5.0\}\), \(d \in \{0.0, 0.33, \ldots, 1.0\}\), and \(e \in \{0.0, 1.0, \ldots, 3.0\}\). We used $\delta=10^{-3}$ for the Huber loss.
\section{Comparing the forms of the scaling laws}
\label{app:comparison_scaling_laws}
\label{model_selection_details}
To evaluate scaling law forms, we employed a form of cross validation which we call step-wise cross-validation. The step-wise cross-validation process involves iterating over various combinations of data thresholds and regularization values, in order to test the ability for the scaling law to predict the performance of models trained on either more pre-training data or on more fine-tuning data points, or both.

\begin{figure}[h]
    \centering
    \includegraphics[width=\textwidth]{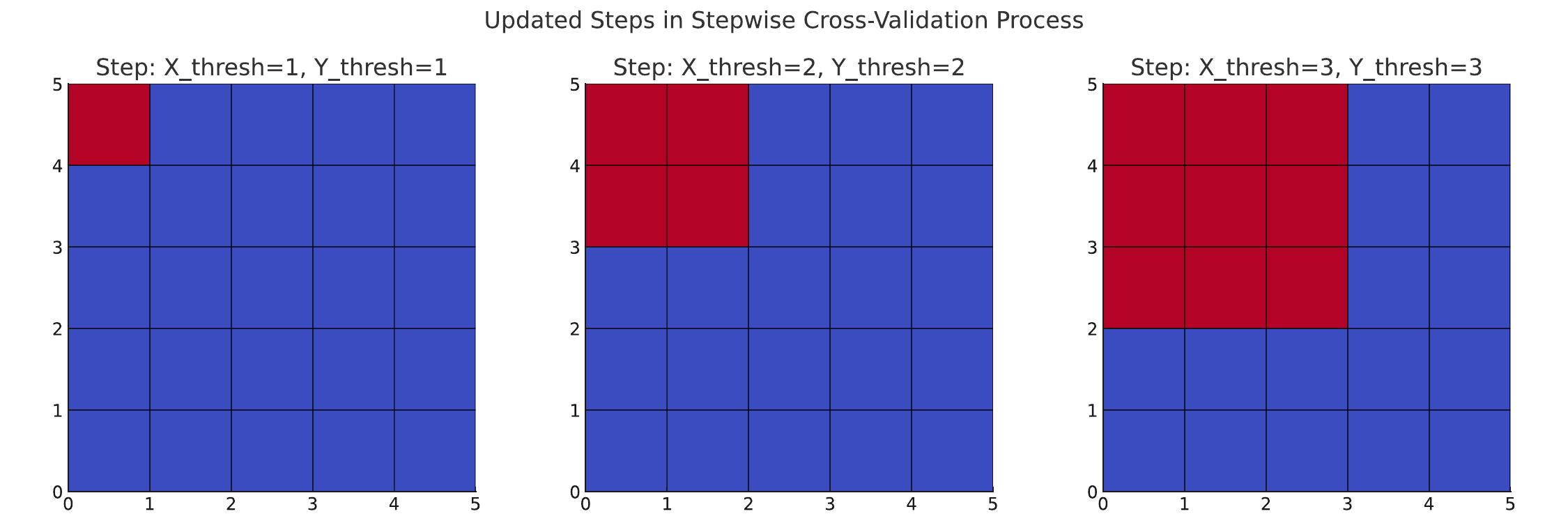}
    \caption{Illustration of selected steps in step-wise cross-validation. The three plots depict key steps: (1) both X and Y thresholds set to 1, (2) both thresholds set to 2, and (3) both thresholds set to 3. The training set (in red) and the validation set (in blue) shift as the thresholds increase. This representation simplifies the actual process, which involves iterating over all possible combinations of X and Y thresholds.}
    \label{fig:Plots/updated_cross_validation}
\end{figure}

This procedure is intended to be a comprehensive form of cross-validation that is less computationally expensive than exhaustive cross-validation, in which all $2^N$ possible combinations of training-test splits are considered. Specifically, the procedure is as follows:
\begin{enumerate}
    \item Threshold Combinations: The dataset is split into training and testing sets based on the threshold combinations. Every \textit{skip\_number}-th combination is used to balance computational efficiency and diversity in data splits.
    \item Training and Testing: A model is trained on the training set and evaluated on the validation set. The \textit{basinhopping} optimization technique is utilized to find the best parameters that minimize the objective function of the scaling law.
    \item L2 Regularization for Exponents (Controlled by \(\alpha\)): This term adds a penalty to the loss function proportional to the sum of squares of the exponent parameters. The regularization strength is controlled by the parameter \(\alpha\), which helps prevent overfitting by penalizing large exponent values.
    \item L2 Regularization for Coefficients (Controlled by \(\beta\)): Similarly, this term penalizes the sum of squares of the coefficient parameters, with \(\beta\) controlling the regularization strength. It ensures that the coefficients do not become excessively large, which can lead to overfitting.
    \item Model Evaluation: The model's performance is assessed using  Root Mean Squared Error (RMSE) and Mean Absolute Error (MAE).
\end{enumerate}
This approach allows for a detailed and thorough investigation of the model's performance across a wide range of scenarios, ensuring robustness and reliability of the parameters chosen for the scaling law function. This approach is summarized in \autoref{fig:Plots/updated_cross_validation}. The following combinations of \(\alpha\) and \(\beta\) were used: $\alpha: [0.0, 0.01, 0.1, 1.0, 5.0, 10, 50]$, $\beta: [0, 0.0001, 0.001, 0.01, 0.1]$. 

\begin{table}[htbp]
\centering
\caption{Scaling laws for transfer with their cross-validation performance on the fictional encyclopedia dataset}
\label{tab:simplified_scaling_laws}
\begin{tabular}{@{}>{\raggedright\arraybackslash}p{1cm}>{\raggedright\arraybackslash}p{8cm}cc@{}}
\toprule
Index & Scaling law & Lowest RSME & Lowest MAE \\ 
\midrule
1 & $L(p, f) = (a_0p^{-a_1}+a_2)f^{-a_3}+a_4$ & 0.126391 & 0.106308 \\
2 & $L(p, f) = (a_0p^{-a_1}+a_2)(f+a_3)^{-a_4}+a_5$ & 0.124945 & 0.110253 \\
3 & $L(p, f) = (a_0(p + a_1)^{-a_2}+a_3)f^{-a_4}+a_5$ & 0.615465 & 0.612981 \\
4 & $L(p, f) = (a_0(p + a_1)^{-a_2}+a_3)(f+a_4)^{-a_5}+a_6$ & 0.534095 & 0.529382 \\
5 & $L(p, f) = (a_0p^{-a_1}+a_2)f^{-a_3}$ & 0.126392 & 0.106309 \\
\bottomrule
\end{tabular}
\end{table}
\section{Analysis of the number of epochs until convergence}
\begin{figure}[htbp]
    \centering
    \begin{subfigure}[b]{0.5\textwidth}
        \centering
        \includegraphics[width=\textwidth]{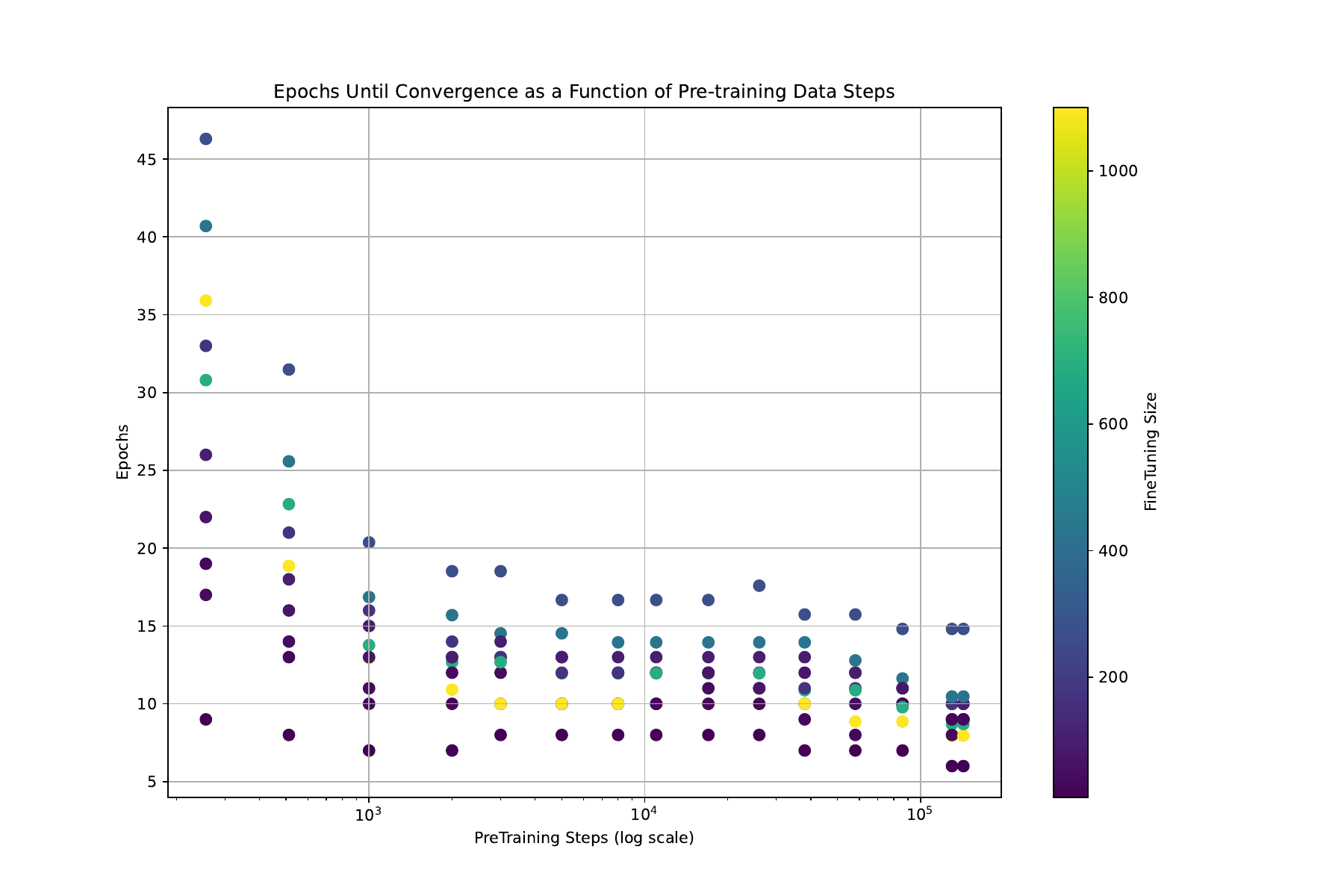} 
        \caption{Epochs as a function of pre-training steps for different fine-tuning data sizes}
        \label{fig:pretraining_epochs}
    \end{subfigure}%
    \hfill 
    \begin{subfigure}[b]{0.5\textwidth}
        \centering
        \includegraphics[width=\textwidth]{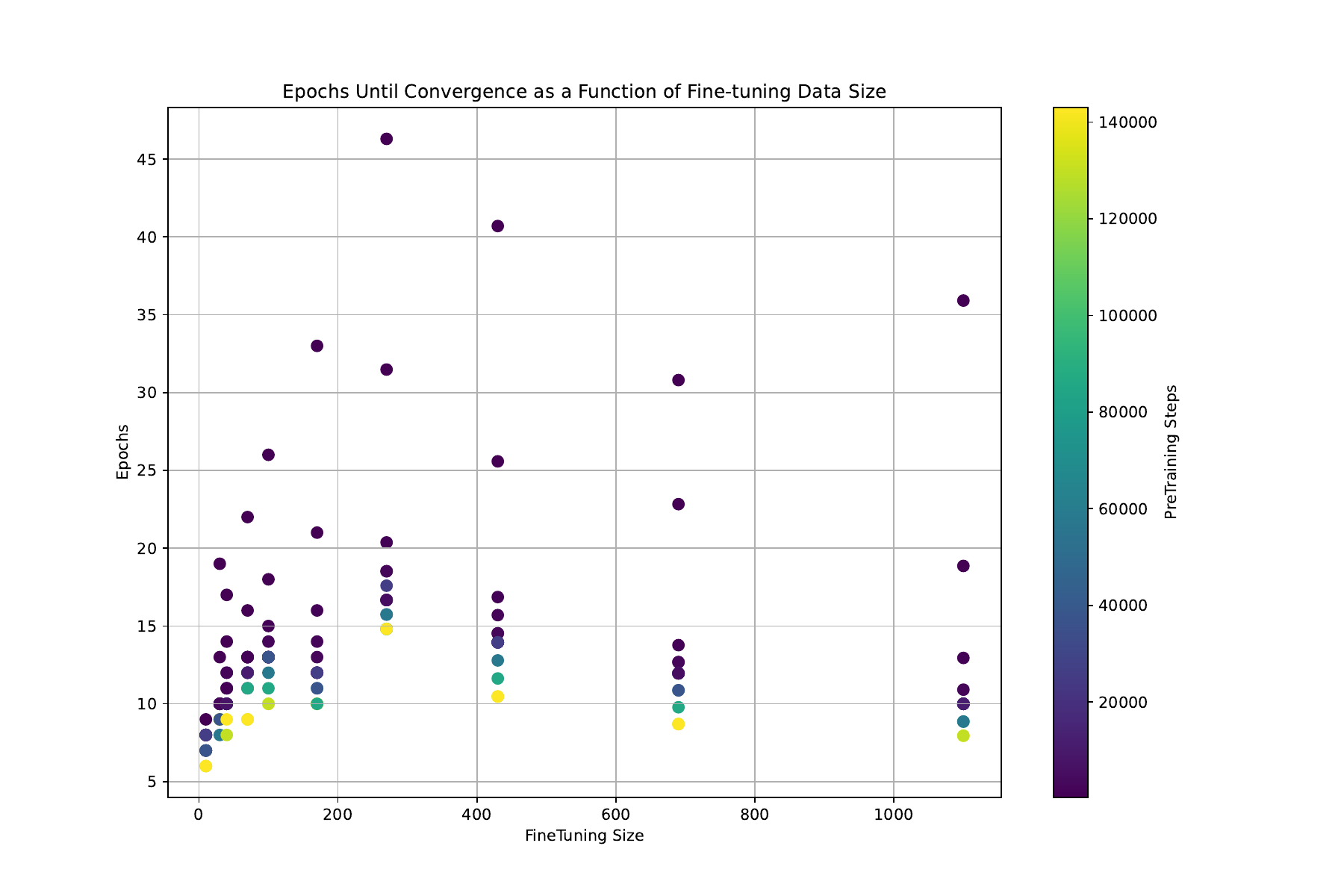} 
        \caption{Epochs as a function of fine-tuning data size segregated by pre-training steps}
        \label{fig:fine_tuning_epochs}
    \end{subfigure}
    \caption{Visual analysis of training epochs variability with respect to pre-training steps and fine-tuning data size. Left: Variation of epochs with pre-training steps across different fine-tuning sizes. Right: Variation of epochs with fine-tuning size, differentiated by pre-training steps.}
    \label{fig:epoch_analysis}
\end{figure}

Similar to \cite{hernandez2021scaling}, we analyzed the behavior of the number of epochs until convergence as a function of training inputs. We found that, just like validation loss, the number of epochs until convergence---according to the early stopping patience of 3---was predictable with pre-training. However, the trend was more unclear for fine-tuning data size. In particular, we found that, as a general rule with a higher degree of pre-training, the model converged in fewer epochs. However, this trend did not clearly hold for fine-tuning data size. These results are summarized by \autoref{fig:epoch_analysis}.

\section{Description and sample from each dataset}
\label{app:dataset_details}
\subsection{Fictional encyclopedia}
This artificial dataset is a fictional encyclopedia of biographical entries in a fantasy world. The fictional encyclopedia was generated from GPT-3.5, which was asked to write biographies for characters inside a universe. Each character had traits generated by a simulation engine written in Python. A sample from this dataset is provided below. The simulation engine was designed to ensure that all of the major details in the universe, such as events that occur within people's lives, would be consistent across biographical entries. The purpose of creating such a training dataset was to provide rich and diverse fine-tuning data that would not be found in the Pile, providing a suitable source to test the structure of scaling laws for transfer for transformer models, without being influenced by text patterns that the model memorized from pre-training. A random entry from the fictional encyclopedia can be found below.

\textbf{Sample from dataset}

\label{sample_fictional_biography}
\begin{mdframed}[backgroundcolor=gray!20, linecolor=blue]
\small
Bel Elur was a prominent figure in the Orilune culture of Elyndorium. Born in the year 16 to parents Ylcar Siljorzor, an Elemental Aligned Chef, and El Ur, a Warlock, Bel's life was destined to be filled with both magic and gastronomy.

From an early age, Bel's potential as a spellcaster was evident. His eloquence and optimistic outlook on life made him a natural spell tutor, and he quickly became known for his profound knowledge and skill in teaching others the ways of magic. However, it was also his cruel streak that set him apart from his peers, and some questioned whether it was a necessary aspect of his nature or a mere result of his upbringing.

Despite the challenges faced by the Orilune culture, Bel was fortunate enough to live in a time of relative stability. However, events such as the local well drying up when he was only a year old posed significant difficulties for his community. Four years later, famine struck, further testing the resilience of the Orilune people.

In the year 25, war broke out between Orilune and the neighboring Korvessi culture. The superior military leadership of the Korvessi led to their victory, leaving the Orilune devastated. This defeat prompted a period of reflection for the Orilune, leading them to strengthen their military forces after a brief trade alliance with the Aelorians in the year 30.

Bel's own life took a turn in the year 47 when the Orilune faced the consequences of a public reading from a forbidden tome. The society plunged into chaos as the forbidden knowledge spread, and it was during this time that Bel's cruel tendencies began to manifest more prominently.

In the year 49, the weakened Orilune military forces faced further setbacks, which left the culture vulnerable to future conflicts. However, in the year 57, a glimmer of hope emerged as a member of the Orilune royalty married into Elandriel's esteemed royal family. This alliance strengthened the ties between the two cultures and offered a sense of security for the Orilune.
\end{mdframed}
\subsection{arXiv math}
This dataset was obtained by querying the most recent papers in the mathematics category of arXiv, and parsing them in latex. Since we queried these papers after the Pythia models were trained, they care unlikely to have been in the pre-training data.

\textbf{Sample from dataset}

\begin{mdframed}[backgroundcolor=gray!20, linecolor=blue]
\small
The purpose of this chapter is to give a rather exhaustive survey on the Maurer--Cartan equation and its related methods, which lie at the core of the present monograph. 
We first give a recollection of the Maurer--Cartan equation and its gauge symmetries in differential geometry.
This chapter is viewed as a motivation for the rest of the book, which consists of higher algebraic generalisations of the key notions of gauge theory. Reading it is not  mandatory to understand what follows but this might help the reader to get some concrete pictures in mind before passing to a more abstract treatment. 
Then, we establish the general theory of the Maurer--Cartan equation in differential graded Lie algebras. 
With that in hand, we discuss the philosophy of deformation theory suggesting that studying Maurer--Cartan elements of differential graded Lie algebras, as well as the symmetries of those elements, is the central question of \emph{any} deformation theory problem in characteristic $0$~. 
In the last chapter~\ref{sec:Applications}, we shall discuss more recent developments making that philosophy precise by means of higher category theory. \\

Throughout this chapter, various infinite series arise. For simplicity, we work with the strong assumption that the various differential graded Lie algebras are nilpotent, so that all these series are actually finite once evaluated on elements. We refer to the treatment of complete algebras given in the next chapter~\ref{sec:OptheoyFilMod} 
for the correct setup in which convergence is understood in the rest of the text.
\end{mdframed}
\subsection{Statistics textbook}
This dataset was taken from Statistics for Ecologists: A Frequentist and Bayesian Treatment of Modern Regression by John Fieberg \cite{fieberg2024statistics}, which is an open source textbook. We found this book from the Open Textbook Library from the University of Minnesota. This textbook was published after the creation of the Pile (\cite{biderman2023pythia}), and therefore we believe it is unlikely to have appeared in the Pile.

\textbf{Sample from dataset}

\begin{mdframed}[backgroundcolor=gray!20, linecolor=blue]
\small
Bootstrap procedure

If our sample data are representative of the population, as we generally assume when we have a large number
of observations selected by simple random sampling, then we can use the distribution of values in our sample
to approximate the distribution of values in the population. For example, we can make many copies of our
sample data and use the resulting data set as an estimate of the whole population. With this estimated
population in place, we could repeatedly sample from it, calculate the statistic for each of these sampled data
sets, and then compute the standard deviation of these sample statistics to form our estimated standard error.
In practice, we do not actually need to make multiple copies of our sample data to estimate the population;
instead, we form new data sets by sampling our sample data with replacement, which effectively does the
same thing. This means that each observation in the original data set can occur zero, one, two, or more times
in the generated data set, whereas it occurred exactly once in the original data set.
This process allows us to determine how much our estimates vary across repeated samples. We quantify this
variability using the standard deviation of our estimates across repeated samples and refer to this standard
deviation as our bootstrap standard error (SE).
• A bootstrap sample is a random sample taken with replacement from the original sample, of the same size 
as the original sample – a “simulated” new sample, in a sense.
• A bootstrap statistic is the statistic of interest, such as the mean or slope of the least-squares regression
line, computed from a bootstrap sample
• A bootstrap distribution is the distribution of bootstrap statistics derived from many bootstrap samples.
Whereas the sampling distribution will typically be centered on the population parameter, the bootstrap
distribution will typically be centered on the sample statistic.2 That is OK, what we really want from a
bootstrap distribution is a measure of variability from sample to sample.
\end{mdframed}

\subsection{House cat genome}
This dataset was taken from the sequenced genome of a house cat, taken from \cite{felis_catus_9_ensemble}.

\textbf{Sample from dataset}

\begin{mdframed}[backgroundcolor=gray!20, linecolor=blue]
\small
\begin{lstlisting}
ATCAGGAGATCTAGATGCCTGGAGAGGAGTGGAGAAAACGGGAAACCCTCTTATGGGAAG
AGGTAATATGTATTTCTCCTTCGAATATAAAAAAAGTAAAAAGAAGGAAAACTTACCAAA
TTCACTTATGAGCCATTCATTACCCTGATACCAAAACCAGATAAAGCCCTCCACTAAAAC
CAAAACTGCAGCGGCGCCTTGTGGGCTCGGTCGGTTTTACTGTCCAACTCTTAATTTCAG
ATTAGGAAATAATCTTGCGGTGCATGGGTTCAAGTCCCACGTTGGACCCTGCCATGACAG
TGTGGGGAATGGCTAGGATTCTCTCTCTCCCTGTCTCTCTGCCCCTCCCTCACTTTTTTG
TACTCTAAGGAAAGAAATAAACATTTAAAAAAATGTTGAAAATTTTTTAAATAAAACTGC
ATACCAATAGCCTTGATGAGTATGTATGCCGAATTCTTCATTAAAAATACCTCAAATTAA
ATTCAAACAATACATTACATAGAATCATTTACCATAATCAAGTGGGATATATCCCTGGCT
TCAAGGGTGGTACCACATTCACAGATCAATCAACATGATGCACCACAGTAATAGAAGAAA
GGATAAGAACAATATGATGCTTTCAACAGATGCAGGAAAACCATTTGACAAAATGTCACA
TCCATTCATGATAAAAACGCTCAGCAAATACATTGAGACTCAACCTACCTGCACATAATA
AAATCAATCTAGGAAAACCCACAGCTAATCTCATCCTCAATGGGGAAAATTGAAAGTTGT
TCCTCCAAGGTCAGGGACAAGACAGGGATGACCCCTCTCACCGACTGTTATTCACATAGT
ACGGGAAATCCTAGCCACAGCAATCAGACAACAAAAAGAAATAAGAGGCATCCAAATCAG
\end{lstlisting}
\end{mdframed}
\subsection{Enron emails}
This dataset was taken from a subset of the Enron Emails dataset, a well-known NLP dataset (\cite{enron_dataset}).

\textbf{Sample from dataset}
\small
\begin{mdframed}[backgroundcolor=gray!20, linecolor=blue]
\texttt{Message-ID: <2450437.1075860190764.JavaMail.evans@thyme>} \\
\texttt{Date: Wed, 14 Feb 2001 02:24:00 -0800 (PST)} \\
\texttt{From: mark.taylor@enron.com} \\
\texttt{Subject: New Canadian EOL Product} \\
\texttt{Mime-Version: 1.0} \\
\texttt{Content-Type: text/plain; charset=us-ascii} \\
\texttt{Content-Transfer-Encoding: 7bit} \\
\texttt{X-From: Mark Taylor} \\
\texttt{X-To: } \\
\texttt{X-cc: } \\
\texttt{X-bcc: } \\
\texttt{X-Folder: \textbackslash Mark\_Taylor\_Jun2001\textbackslash Notes Folders\textbackslash All documents} \\
\texttt{X-Origin: Taylor-M} \\
\texttt{X-FileName: mtaylor.nsf} \\
\\
\texttt{Greg:} \\
\\
\texttt{2 Issues - an easy language issue and a more complicated regulatory issue.} \\
\\
\texttt{First the easy one.}
\end{mdframed}
\end{document}